\documentclass[onecolumn]{main}

\paperstyle{fancy}
\papercolor{blue}
\usepackage{float}
\usepackage{booktabs}
\usepackage{graphicx}

\title{EEGAgentBench: Benchmarking LLM Agents on Short- and Long-Horizon EEG Analysis}
\author[1,2]{Huyu Wu}
\author[1,2]{Weining Weng}
\author[1,2]{Yuchen Liu}
\author[1,2]{Yiqiang Chen}
\author[1,2,\dagger]{Yang Gu}

\affiliation[1]{Institute of Computing Technology, Chinese Academy of Sciences, Beijing, China}
\affiliation[2]{University of Chinese Academy of Sciences, Beijing, China}
\contribution[\dagger]{Corresponding author}
\makeatletter
\gappto\sa@contributionlist{%
  \par\vspace{0.08cm}
  {\centering\normalfont\href{mailto:wuhuyu25s@ict.ac.cn, wengweining21b@ict.ac.cn, liuyuchen23s@ict.ac.cn, yqchen@ict.ac.cn, guyang@ict.ac.cn}{\textcolor{metalink}{\texttt{\{wuhuyu25s, wengweining21b, liuyuchen23s, yqchen, guyang\}@ict.ac.cn}}}\par}
  \vspace{0.25cm}
  {\raggedright\normalfont\small\sffamily\bfseries
    \href{https://github.com/528why/EEGAgentBenchmark}{\textcolor{metalink}{\faGithub\hspace{0.35em}Code}}\hspace{1.25em}
    \href{https://528why.github.io/EEGAgentBench-Page/}{\textcolor{metalink}{\faGlobe\hspace{0.35em}Project}}\hspace{1.25em}
    \href{https://huggingface.co/datasets/whhhy123/EEGAgentBenchmark}{\textcolor{metalink}{\faDatabase\hspace{0.35em}Dataset}}\par}
  \vspace{-0.15cm}
}
\makeatother

\abstract{
Electroencephalography (EEG) analysis is evolving from short-segment classification toward long-horizon interpretation that demands iterative evidence accumulation, multi-step reasoning, and coordinated use of specialized signal-processing tools. Although large language models (LLMs) have recently shown promise as autonomous agents for EEG analysis, existing EEG agentic evaluations remain fragmented, covering limited tasks over narrow temporal horizons with inconsistent protocols, and providing no comprehensive assessment of agents' reasoning, tool-use, and workflow construction capabilities. To address this gap, we propose \textbf{EEGAgentBench}, a unified benchmark for systematically evaluating LLM agents on short- and long-horizon EEG analysis. EEGAgentBench spans six representative EEG applications ranging from knowledge question answering to sleep staging. It encompasses signal durations from 2 seconds to nearly 23 hours, with prediction targets ranging from class labels to event intervals and epoch-level sequences. This design supports unified evaluation across knowledge reasoning, short-horizon interpretation, long-horizon event detection, and sequential understanding. The benchmark further provides 10 deterministic EEG analysis tools that expose only task-relevant signal measurements. Agents must therefore select tools autonomously, accumulate evidence iteratively, and construct multi-step workflows. For evaluation, we benchmark 29 frontier LLMs from 15 model families. Results demonstrate that EEGAgentBench effectively distinguishes agent capabilities beyond model scale and inference cost, while revealing substantial limitations of current LLM agents in long-horizon EEG analysis, particularly in sustained evidence accumulation and multi-step reasoning.
}
 
\usepackage{listings}
\usepackage{tikz}

\newcommand{\tool}[1]{\texttt{\detokenize{#1}}}
\definecolor{bestcolor}{HTML}{4169E1}
\definecolor{highlightrow}{HTML}{E8F4FD}
\definecolor{goldmedal}{HTML}{D4AF37}
\definecolor{silvermedal}{HTML}{B7BDC8}
\definecolor{bronzemedal}{HTML}{CD7F32}

\newcommand{\scorebold}[1]{\textbf{#1}}
\newcommand{\best}[1]{{\color{bestcolor}{#1}}}
\newcommand{\bestscore}[1]{\best{\scorebold{#1}}}
\newcommand{\secondscore}[1]{\textit{#1}}
\newcommand{\rankmedal}[2]{\tikz[baseline=-0.65ex,scale=0.19]{\fill[bestcolor] (-0.58,1.18) -- (-0.10,0.40) -- (-0.70,0.40) -- cycle;
    \fill[red!75!black] (0.58,1.18) -- (0.10,0.40) -- (0.70,0.40) -- cycle;
    \filldraw[fill=#1,draw=black!35,line width=0.25pt] (0,0) circle (0.70);
    \node[font=\bfseries\tiny,inner sep=0pt] at (0,0) {#2};
  }}
\newcommand{\goldrank}{\rankmedal{goldmedal}{1}}
\newcommand{\silverrank}{\rankmedal{silvermedal}{2}}
\newcommand{\bronzerank}{\rankmedal{bronzemedal}{3}}

\definecolor{cRose}{HTML}{CD708D}
\definecolor{cCream}{HTML}{F0E3D1}
\definecolor{cGrey}{HTML}{B6AFA6}
\definecolor{cBrown}{HTML}{50362E}
\definecolor{cPlum}{HTML}{291A42}
\definecolor{cRasp}{HTML}{C85170}
\lstdefinestyle{traj}{basicstyle=\ttfamily\scriptsize\color{cBrown},
  backgroundcolor=\color{cCream!40},
  breaklines=true, breakatwhitespace=false, breakindent=0pt,
  columns=fullflexible, keepspaces=true, showstringspaces=false, upquote=true,
  frame=single, rulecolor=\color{cBrown!60}, framesep=4pt, xleftmargin=4pt,
  aboveskip=5pt, belowskip=3pt,
  moredelim=[l][\color{cRasp}\bfseries]{----},
  moredelim=[l][\color{cPlum}\itshape]{[Thinking]},
  moredelim=[l][\color{cRasp}\bfseries]{[Tool]},
  moredelim=[l][\color{cGrey!45!black}]{[Result]},
  moredelim=[l][\color{cBrown}\bfseries]{INPUT:},
  moredelim=[l][\color{cGrey!45!black}]{GROUND},
  moredelim=[l][\color{cRasp}\bfseries]{FINAL},
  moredelim=[l][\color{cRasp}\bfseries]{SCORE:},
}
\lstdefinestyle{inbox}{basicstyle=\ttfamily\scriptsize\color{cBrown},
  backgroundcolor=\color{cCream!45},
  breaklines=true, columns=fullflexible, keepspaces=true,
  showstringspaces=false, upquote=true,
  frame=single, rulecolor=\color{cBrown!55}, framesep=4pt, xleftmargin=3pt,
  aboveskip=6pt, belowskip=2pt,
  moredelim=[l][\color{cRasp}\bfseries]{==},
}
\lstdefinestyle{inboxcompact}{style=inbox,
  basicstyle=\ttfamily\fontsize{6.4pt}{7.1pt}\selectfont\color{cBrown},
  framesep=3pt,
  aboveskip=4pt, belowskip=1pt,
  emptylines=0,
}

\begin{document}
\enlargethispage{0.65cm}
\vspace*{-2.05cm}
\noindent
\begin{tikzpicture}
  \draw[draw=rulecolor,line width=0.9pt]
    (0.5pt,0) -- (\dimexpr\linewidth-0.5pt\relax,0);
  \node[anchor=south west,inner sep=0pt] at (0,0.05cm)
    {\includegraphics[height=1.35cm]{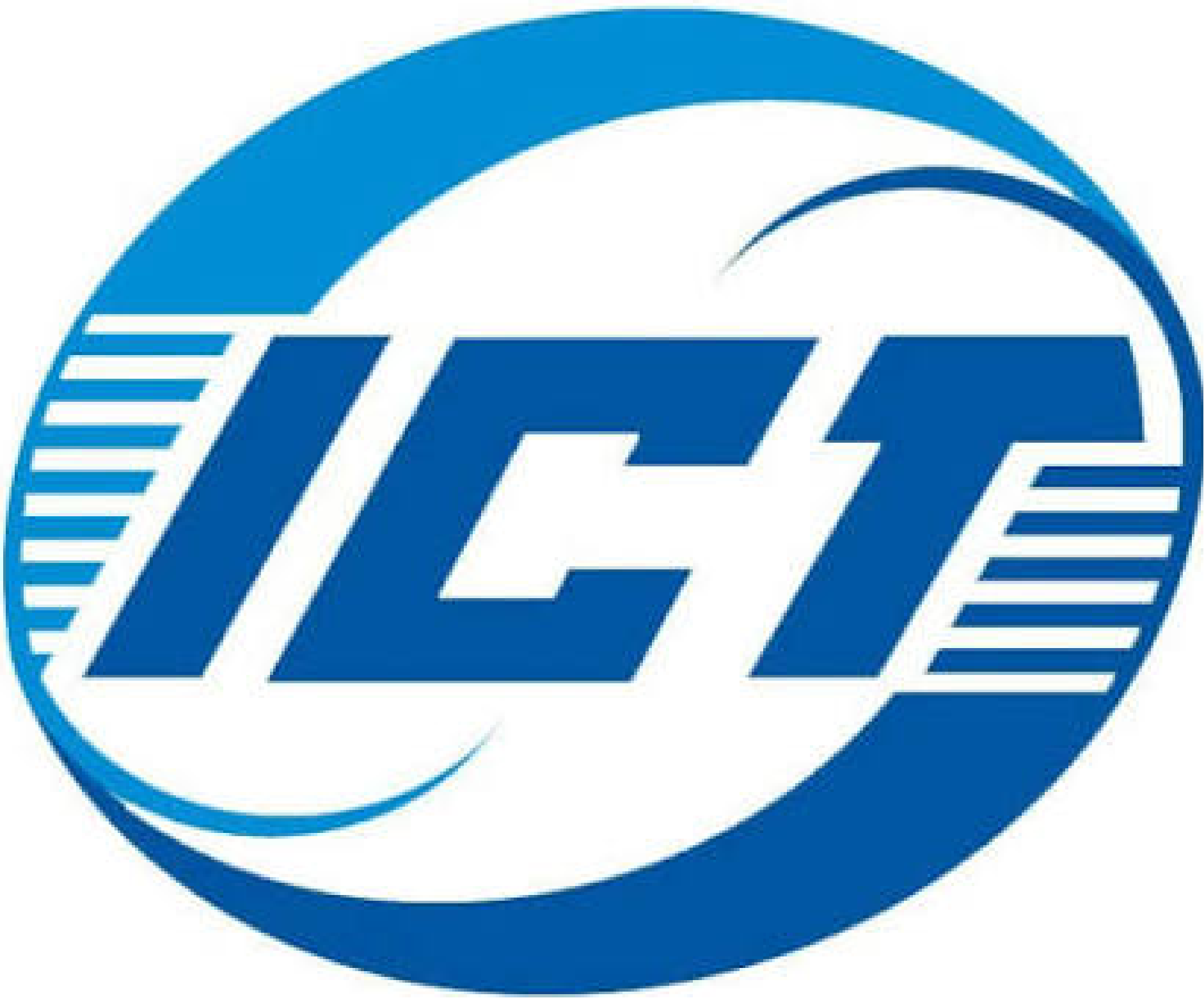}};
\end{tikzpicture}
\par\vspace{0.30cm}
\maketitle

\begin{center}
  \includegraphics[width=0.95\linewidth]{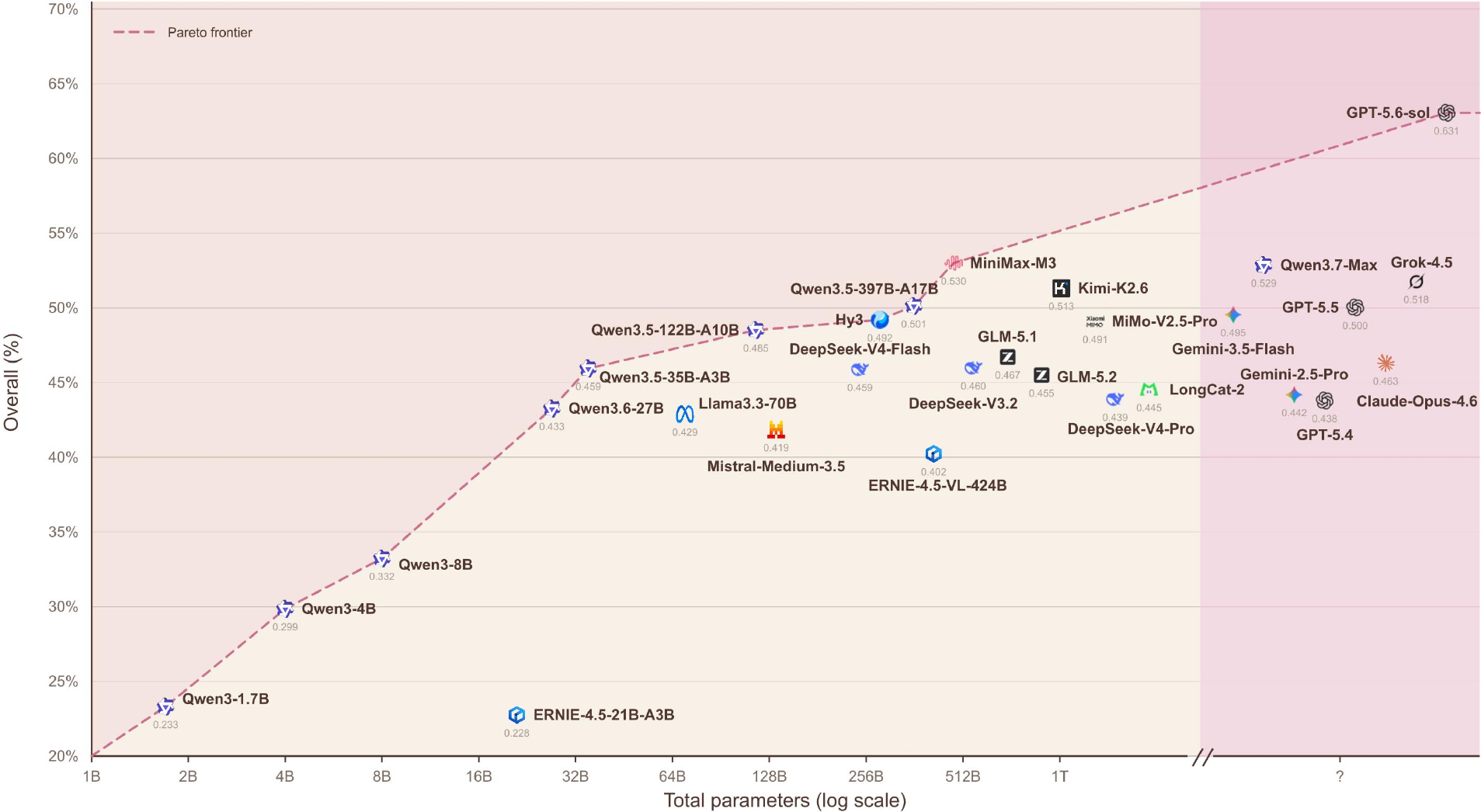}
  \captionof{figure}{Model scale versus performance.}
  \label{fig:model_size_performance}
\end{center}

\newpage
\section{Introduction}
Electroencephalography (EEG) records brain activity as multichannel time-series signals at millisecond resolution, serving as a fundamental tool in neuroscience and neurology~\cite{subha2010eeg}. Computational EEG analysis has long relied on task-specific models. Recent advances in deep learning and EEG foundation models have substantially improved seizure detection, sleep staging, artifact identification, and neurological disorder screening~\citep{DBLP:journals/corr/abs-1901-05498,lawhern2018eegnet,perslev2021u,jiang2024large,wang2024eegpt,wang2025cbramod,jiang2025neurolm}. However, most existing EEG systems still work as predictors: they process predefined signal segments through fixed analysis pipelines to produce task-specific outputs. This paradigm presupposes that the evidence and analysis procedures are predefined, not autonomously selected from the recording's evolving properties.

Recent advances in LLMs have enabled agentic EEG analysis, allowing models to reason over signals, invoke specialized tools, and construct adaptive analysis workflows. EEGAgent employs an LLM to orchestrate preprocessing, feature extraction, event detection, and report generation~\citep{zhao2025eegagent}. BrainAgent further organizes sleep and emotion analysis through a supervisor and specialized sub-agents with hierarchical evaluation from atomic operations to long workflows~\citep{zhou2026brainagent}. These studies demonstrate the feasibility of LLM-based EEG agents while motivating systematic evaluation across diverse tasks and temporal horizons.

However, existing EEG agent evaluations remain fragmented, with limited tasks, narrow temporal horizons, and no systematic assessment of reasoning, tool use, or workflow construction. As EEG analysis moves from short-segment prediction to long-horizon interpretation, evaluations must assess extended-recording analysis, context-spanning evidence accumulation, adaptive tool invocation, and iterative decision refinement. Current benchmarks do not meet these requirements. EEGAgent relies mainly on case studies and predefined models~\citep{zhao2025eegagent}. BrainAgent offers a structured framework but includes only 60 instructions focused on sleep and emotion~\citep{zhou2026brainagent}. Traditional EEG benchmarks cover broader tasks, yet target predictive models under fixed pipelines rather than autonomous agents~\citep{jayaram2018moabb,kastrati2025eeg,banville2026neuralbench}. This leaves one fundamental question unanswered:

\vspace{0.75\baselineskip}
\noindent\textbf{\textit{Can LLM agents autonomously reason over EEG signals, select appropriate signal-processing tools, and construct effective analysis workflows across diverse tasks?}}
\par\vspace{0.75\baselineskip}

We introduce EEGAgentBench, a unified benchmark for evaluating general-purpose LLM agents on tool-based EEG analysis. EEGAgentBench comprises 1,072 evaluation instances from six public datasets, spanning six representative EEG tasks across knowledge reasoning, short-horizon signal interpretation, and long-horizon event and sequence analysis. It covers recordings ranging from 2 seconds to nearly 23 hours, prediction targets from classification to event localization and sequential labeling. The benchmark also provides a standardized toolbox of 10 deterministic, non-parametric EEG analysis tools. Rather than following predefined pipelines, agents must autonomously select tools, iteratively accumulate evidence, and construct task-specific analysis workflows. We benchmark 29 frontier LLMs from 15 model families, including both open-weight and proprietary models, and systematically analyze task performance, tool use, computational efficiency, and long-horizon failure modes. Results demonstrate that EEGAgentBench effectively distinguishes agent capabilities beyond model scale and inference cost, while revealing that current LLM agents still struggle with long-horizon EEG analysis, exposing fundamental limitations in sustained evidence accumulation and multi-step reasoning. Our contributions are summarized as follows:

\begin{itemize}
\item We introduce \textbf{EEGAgentBench}, the first unified benchmark that systematically evaluates general-purpose LLM agents on both short- and long-horizon EEG analysis using standardized tool-based workflows.

\item We design six representative EEG tasks and a deterministic toolbox of ten non-parametric signal-processing tools, which together unify the assessment of knowledge reasoning, adaptive tool use, and workflow construction across diverse EEG applications.

\item We evaluate 29 frontier LLMs at scale and comprehensively examine their task performance, tool use behavior, computational efficiency, and long-horizon failure modes, establishing strong baselines and revealing key limitations of current EEG agents.
\end{itemize}

\section{Related Work}
\label{sec:related_work}

\subsection{EEG Analysis Methods}

Computational EEG analysis has evolved from hand-crafted feature engineering to deep learning and, more recently, EEG foundation models~\citep{DBLP:journals/corr/abs-1901-05498,lawhern2018eegnet,perslev2021u,jiang2024large,wang2024eegpt,wang2025cbramod,jiang2025neurolm}. These methods improve performance and cross-task generalization. However, they remain predictor-oriented. They map predefined EEG recordings directly to task outputs. Recent studies extend this paradigm to LLM agents. EEGAgent employs an LLM to orchestrate EEG processing tools, while BrainAgent further introduces a hierarchical multi-agent framework for sleep and emotion analysis~\citep{zhao2025eegagent,zhou2026brainagent}. Together, they demonstrate the feasibility of tool-assisted EEG agents but lack systematic evaluation across diverse EEG tasks.

\subsection{Benchmarking EEG Analysis}

Existing EEG benchmarks evaluate predictive models under standardized datasets, protocols, and metrics. MOABB focuses on brain-computer interface tasks~\citep{jayaram2018moabb}, while EEG-Bench and NeuralBench extend evaluation to broader clinical and neuroscience applications, including EEG foundation models~\citep{kastrati2025eeg,banville2026neuralbench}. However, these benchmarks assume predefined prediction pipelines and are not designed for interactive LLM agents. Current EEG-agent evaluations remain limited to capability demonstrations or small-scale instruction sets~\citep{zhao2025eegagent,zhou2026brainagent}, leaving no unified benchmark for systematically assessing agent capabilities.

\subsection{Interactive Evaluation of LLM Agents}

Recent agentic benchmarks evaluate agents through interactions with external environments rather than single-response generation. AgentBench studies decision making across interactive environments~\citep{liu2024agentbench}; GAIA evaluates retrieval, reasoning, and tool use on realistic tasks~\citep{mialon2024gaia}; and $\tau$-bench verifies task completion through executable environments and APIs~\citep{yao2024tau}. AstaBench emphasizes reproducible tool access and analyzes execution behavior together with computational cost~\citep{bragg2025astabench}. These benchmarks highlight the importance of standardized interactive environments, verifiable task outcomes, and systematic analysis of tool use and execution efficiency.

\section{EEGAgentBench}

\begin{figure}[!t]
\centering
\includegraphics[width=0.9\linewidth]{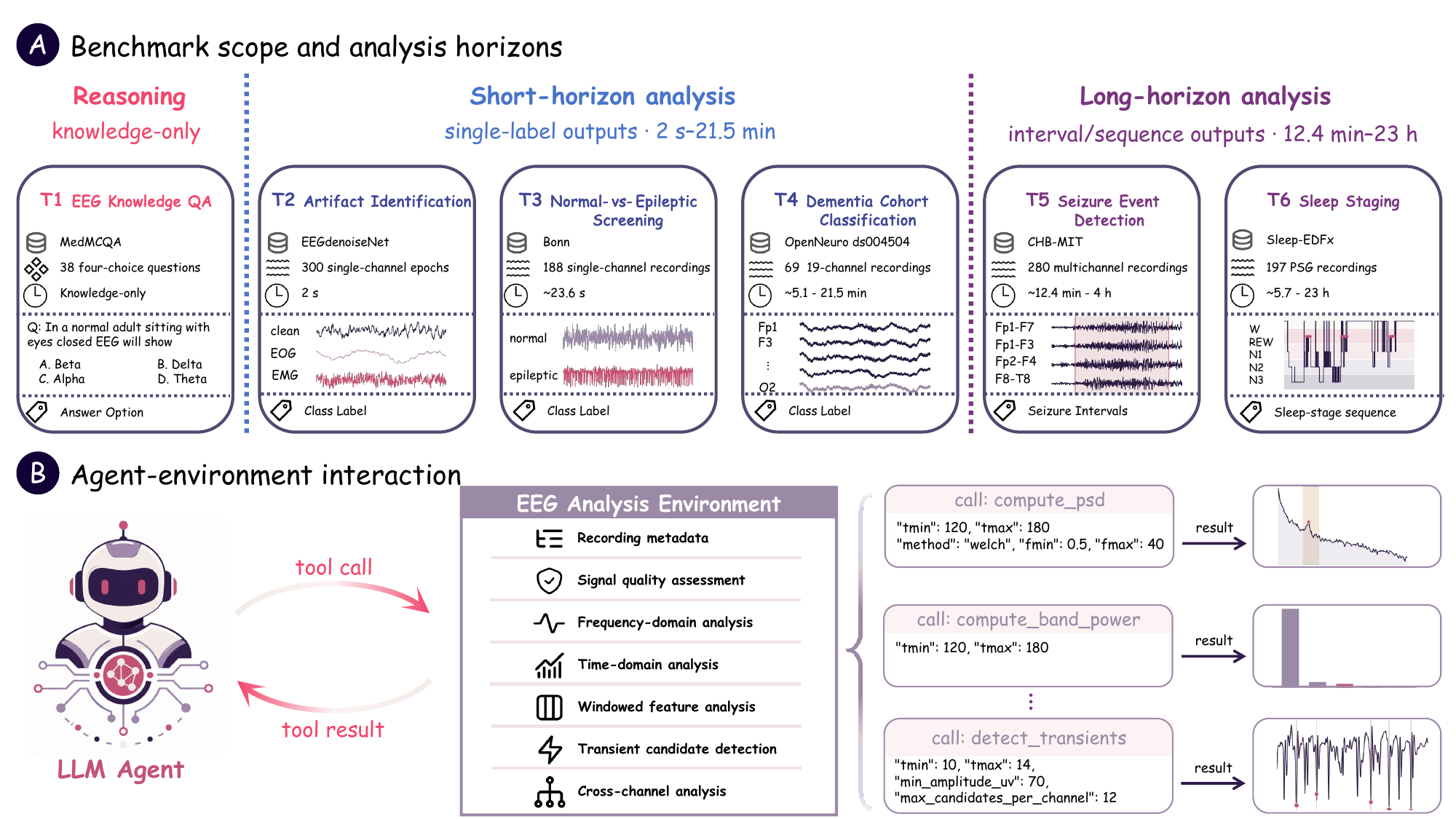}
\caption{Overview of the six EEGAgentBench tasks and the agent--environment interaction.}
\label{fig:benchmark_overview}
\end{figure}

Innovatively, we design EEGAgentBench, a unified benchmark for evaluating LLM agents on EEG knowledge reasoning and tool-assisted signal analysis. As illustrated in Figure~\ref{fig:benchmark_overview}, EEGAgentBench consists of two components: (1) six benchmark tasks covering knowledge reasoning, short-horizon analysis, and long-horizon analysis; (2) an interactive EEG analysis environment with deterministic signal-processing tools that enables autonomous reasoning and adaptive tool use.

\subsection{Benchmark Tasks and Evaluation Metrics}
\label{sec:benchmark_tasks}

\subsubsection{Task Design Philosophy}
\label{sec:task_design_philosophy}
EEGAgentBench follows recent holistic benchmark design principles that emphasize \textbf{capability coverage}. These principles advocate assessing systems through representative tasks that collectively capture the major capabilities required in real-world applications rather than relying on isolated tasks or aggregate scores~\citep{HHLE,MMMU}. Accordingly, the benchmark aims to characterize agent capabilities across diverse EEG analysis scenarios instead of maximizing performance on individual tasks.

Guided by this principle, we organize EEGAgentBench into three complementary capability dimensions: 
\begin{itemize}
    \item \textbf{EEG knowledge reasoning}, which evaluates prerequisite domain knowledge
    \item \textbf{Short-horizon analysis}, which assesses the interpretation of localized EEG patterns and diagnostic evidence
    \item \textbf{Long-horizon analysis}, which measures the ability to accumulate evidence, maintain analysis state, and reason over extended recordings
\end{itemize}

These three capability dimensions are instantiated through six representative benchmark tasks, providing a comprehensive yet structured evaluation across the diverse analytical requirements. \textbf{Knowledge reasoning} contains T1 EEG Knowledge QA, in which a model answers four-choice medical questions related to EEG. \textbf{Short-horizon analysis} contains T2 artifact contamination identification, T3 normal-vs-epileptic screening, and T4 dementia cohort classification. These tasks ask the agent to identify signal contamination, distinguish normal from epileptic activity, and classify research cohorts as Alzheimer's disease (AD), frontotemporal dementia (FTD), or healthy controls (HC). The T4 task is a research cohort classification task rather than a clinical diagnosis task. \textbf{Long-horizon analysis} contains T5 seizure event detection and T6 sleep staging. T5 asks the agent to identify a variable number of seizure intervals in recordings up to about 4 hours. T6 asks the agent to generate a five-class sleep-stage sequence for polysomnography recordings of up to nearly 23 hours.

We distinguish short- and long-horizon analysis by three criteria: the temporal coverage of relevant evidence, the structure and granularity of the required output, and the analysis state maintained across tool interactions. T2--T4 are short-horizon tasks, where agents summarize recordings from 2 seconds to 21.5 minutes into a single prediction based on localized evidence. In contrast, T5 requires identifying zero or more seizure events from recordings spanning 12.4 minutes to 4 hours, while T6 requires generating complete sleep-stage sequences from PSG recordings lasting 5.7 to 23 hours, comprising 681--2,758 consecutive 30-second intervals. Compared with short-horizon tasks, these settings require sustained evidence accumulation, persistent analysis state, and temporally structured outputs rather than a single prediction.

\begin{table}[H]
\centering
\caption{Overview of the six EEGAgentBench tasks. PSG denotes polysomnography.}
\label{tab:tasks}
\scriptsize
\resizebox{\textwidth}{!}{%
\begin{tabular}{@{}lllllll@{}}
\toprule
Setting & Task & Data source & $N$ & Signal scale & Output & Main metrics \\
\midrule
Knowledge     & T1 EEG Knowledge QA       & MedMCQA-EEG-strict & 38  & No signal                              & One label (A/B/C/D)                & Accuracy \\
Short horizon & T2 Artifact Identification & EEGdenoiseNet     & 300 & 2 s, 1 channel                         & One label (clean/ocular/muscle)    & Accuracy, Macro-F1 \\
Short horizon & T3 Normal-vs-Epileptic    & Bonn               & 188 & 23.6 s, 1 channel                      & One label (normal/epileptic)       & Accuracy, Macro-F1 \\
Short horizon & T4 Dementia Cohort        & ds004504           & 69  & 5.1--21.5 min, 19 channels             & One label (AD/FTD/HC)              & Accuracy, Macro-F1 \\
Long horizon  & T5 Seizure Detection      & CHB-MIT            & 280 & 12.4 min--4.0 h, 18--32 EEG channels   & Event interval list                & Dice-S, Event-F1 \\
Long horizon  & T6 Sleep Staging          & Sleep-EDFx         & 197 & 5.7--23.0 h, 5--7 PSG channels         & 30-s epoch label sequence          & Macro-F1, Cohen's kappa \\
\bottomrule
\end{tabular}}
\end{table}

Table~\ref{tab:tasks} summarizes data size, input range, output format, and metrics of each task. Every task is scored automatically from final output. Full metric definitions and rules for malformed outputs are provided in Sections~\ref{app:metrics}--\ref{app:dice_seizure}.

\begin{table}[!t]
\caption{\label{tab:main_results}Main results on EEGAgentBench. Acc. denotes accuracy, Event-F1 denotes the subject-averaged SzCORE-style event-based F1~\citep{DBLP:journals/corr/abs-2402-13005}, and Dice-S denotes Dice@seizure. Overall is the unweighted mean of T1 accuracy, Macro-F1 for T2--T4 and T6, and Event-F1 for T5. Best values are shown in \bestscore{blue bold}, and second-best values are \secondscore{italicized}.}
\centering
\setlength{\tabcolsep}{2pt}
\renewcommand{\arraystretch}{1.05}
\resizebox{\textwidth}{!}{\begin{tabular}{c l *{12}{c}}
\toprule
\multirow{2}{*}{\textbf{Rank}} & \multirow{2}{*}{\textbf{Model}} & \multirow{2}{*}{\textbf{Overall}$\uparrow$} & \multicolumn{1}{c}{T1 Knowledge} & \multicolumn{2}{c}{T2 Artifact} & \multicolumn{2}{c}{T3 Screening} & \multicolumn{2}{c}{T4 Cohort} & \multicolumn{2}{c}{T5 Seizure} & \multicolumn{2}{c}{T6 Sleep} \\
\cmidrule(lr){4-4} \cmidrule(lr){5-6} \cmidrule(lr){7-8} \cmidrule(lr){9-10} \cmidrule(lr){11-12} \cmidrule(lr){13-14}
 & & & Acc.$\uparrow$ & F1$\uparrow$ & Acc.$\uparrow$ & F1$\uparrow$ & Acc.$\uparrow$ & F1$\uparrow$ & Acc.$\uparrow$ & Event-F1$\uparrow$ & Dice-S$\uparrow$ & F1$\uparrow$ & $\kappa\uparrow$ \\
\midrule
\addlinespace[2pt]
\rowcolor{highlightrow}
\multicolumn{14}{l}{\textcolor{gray}{\small\textit{Proprietary models}}} \\
\goldrank & GPT-5.6-sol & \bestscore{0.631} & 0.816 & 0.599 & 0.643 & \secondscore{0.898} & \secondscore{0.899} & \bestscore{0.429} & \bestscore{0.493} & \bestscore{0.549} & \secondscore{0.404} & \bestscore{0.492} & \bestscore{0.488} \\
4 & Grok-4.5 & 0.518 & 0.789 & 0.563 & 0.637 & 0.620 & 0.654 & 0.239 & 0.348 & 0.522 & 0.367 & \secondscore{0.372} & \secondscore{0.277} \\
13 & Claude-Opus-4.6 & 0.463 & \secondscore{0.842} & 0.498 & 0.567 & 0.543 & 0.612 & 0.294 & 0.362 & 0.289 & 0.366 & 0.312 & 0.204 \\
7 & GPT-5.5 & 0.500 & 0.737 & 0.651 & 0.690 & 0.543 & 0.612 & 0.194 & 0.333 & \secondscore{0.530} & \bestscore{0.449} & 0.347 & 0.263 \\
21 & GPT-5.4 & 0.438 & 0.684 & 0.730 & 0.733 & 0.790 & 0.798 & 0.205 & 0.319 & 0.069 & 0.063 & 0.150 & -0.004 \\
19 & Gemini-2.5-Pro & 0.442 & 0.684 & 0.576 & 0.610 & 0.485 & 0.569 & 0.279 & 0.290 & 0.291 & 0.299 & 0.336 & 0.252 \\
\bronzerank & Qwen3.7-Max & 0.529 & 0.763 & 0.747 & 0.770 & \bestscore{0.909} & \bestscore{0.910} & 0.208 & 0.319 & 0.325 & 0.305 & 0.219 & 0.094 \\
8 & Gemini-3.5-Flash & 0.495 & 0.684 & 0.536 & 0.590 & 0.709 & 0.723 & 0.376 & \secondscore{0.464} & 0.377 & 0.378 & 0.291 & 0.193 \\
\addlinespace[2pt]
\rowcolor{highlightrow}
\multicolumn{14}{l}{\textcolor{gray}{\small\textit{Open-weight models: above 1T parameters}}} \\
18 & LongCat-2 & 0.445 & 0.737 & 0.532 & 0.557 & 0.724 & 0.739 & 0.255 & 0.348 & 0.242 & 0.250 & 0.182 & 0.054 \\
20 & DeepSeek-V4-Pro & 0.439 & 0.684 & 0.560 & 0.640 & 0.726 & 0.734 & 0.194 & 0.333 & 0.275 & 0.269 & 0.194 & 0.081 \\
10 & MiMo-V2.5-Pro & 0.491 & 0.763 & 0.745 & 0.747 & 0.624 & 0.660 & 0.281 & 0.391 & 0.313 & 0.261 & 0.218 & 0.065 \\
5 & Kimi-K2.6 & 0.513 & \secondscore{0.842} & 0.642 & 0.670 & 0.767 & 0.777 & 0.272 & 0.319 & 0.347 & 0.341 & 0.208 & 0.071 \\
\addlinespace[2pt]
\rowcolor{highlightrow}
\multicolumn{14}{l}{\textcolor{gray}{\small\textit{Open-weight models: 100B--1T parameters}}} \\
17 & GLM-5.2 & 0.455 & 0.816 & 0.629 & 0.700 & 0.490 & 0.580 & 0.163 & 0.319 & 0.329 & 0.357 & 0.304 & 0.205 \\
12 & GLM-5.1 & 0.467 & 0.763 & 0.636 & 0.670 & 0.704 & 0.723 & 0.198 & 0.348 & 0.265 & 0.320 & 0.236 & 0.114 \\
14 & DeepSeek-V3.2 & 0.460 & 0.711 & 0.631 & 0.693 & 0.776 & 0.777 & 0.163 & 0.304 & 0.271 & 0.258 & 0.207 & 0.075 \\
\silverrank & MiniMax-M3 & \secondscore{0.530} & 0.816 & 0.666 & 0.683 & 0.754 & 0.755 & 0.360 & 0.420 & 0.354 & 0.270 & 0.229 & 0.089 \\
25 & ERNIE-4.5-VL-424B & 0.402 & 0.737 & 0.451 & 0.457 & 0.654 & 0.654 & \secondscore{0.391} & 0.449 & 0.034 & 0.019 & 0.146 & 0.037 \\
6 & Qwen3.5-397B-A17B & 0.501 & 0.763 & \bestscore{0.786} & \bestscore{0.783} & 0.711 & 0.718 & 0.193 & 0.333 & 0.370 & 0.321 & 0.185 & 0.037 \\
9 & Hy3 & 0.492 & 0.763 & \secondscore{0.774} & \secondscore{0.777} & 0.608 & 0.649 & 0.307 & 0.391 & 0.236 & 0.243 & 0.264 & 0.173 \\
16 & DeepSeek-V4-Flash & 0.459 & 0.763 & 0.495 & 0.587 & 0.805 & 0.803 & 0.222 & 0.348 & 0.284 & 0.320 & 0.185 & 0.047 \\
24 & Mistral-Medium-3.5 & 0.419 & 0.737 & 0.505 & 0.580 & 0.568 & 0.628 & 0.220 & 0.348 & 0.264 & 0.274 & 0.217 & 0.079 \\
11 & Qwen3.5-122B-A10B & 0.485 & 0.711 & 0.701 & 0.710 & 0.656 & 0.676 & 0.338 & 0.420 & 0.298 & 0.250 & 0.208 & 0.067 \\
\addlinespace[2pt]
\rowcolor{highlightrow}
\multicolumn{14}{l}{\textcolor{gray}{\small\textit{Open-weight models: 10--100B parameters}}} \\
23 & Llama3.3-70B & 0.429 & \bestscore{0.868} & 0.452 & 0.470 & 0.653 & 0.654 & 0.324 & 0.406 & 0.104 & 0.016 & 0.173 & 0.023 \\
15 & Qwen3.5-35B-A3B & 0.459 & 0.711 & 0.622 & 0.640 & 0.702 & 0.702 & 0.313 & 0.391 & 0.248 & 0.248 & 0.161 & 0.035 \\
22 & Qwen3.6-27B & 0.433 & 0.632 & 0.542 & 0.653 & 0.713 & 0.718 & 0.232 & 0.348 & 0.294 & 0.304 & 0.184 & 0.059 \\
29 & ERNIE-4.5-21B-A3B & 0.228 & 0.526 & 0.167 & 0.333 & 0.301 & 0.394 & 0.220 & 0.217 & 0.000 & 0.000 & 0.151 & 0.010 \\
\addlinespace[2pt]
\rowcolor{highlightrow}
\multicolumn{14}{l}{\textcolor{gray}{\small\textit{Open-weight models: below 10B parameters}}} \\
26 & Qwen3-8B & 0.332 & 0.632 & 0.221 & 0.347 & 0.488 & 0.521 & 0.287 & 0.362 & 0.143 & 0.036 & 0.224 & 0.081 \\
27 & Qwen3-4B & 0.299 & 0.395 & 0.405 & 0.417 & 0.509 & 0.548 & 0.248 & 0.304 & 0.047 & 0.014 & 0.188 & 0.065 \\
28 & Qwen3-1.7B & 0.233 & 0.421 & 0.261 & 0.347 & 0.345 & 0.505 & 0.241 & 0.319 & 0.000 & 0.000 & 0.131 & -0.040 \\
\bottomrule
\end{tabular}}
\end{table}

\subsection{Interactive Environment and EEG Tools}

EEGAgentBench formulates EEG analysis as an interactive process between an agent and an EEG environment, requiring agents to iteratively acquire evidence, reason over observations, and produce a final answer.

To support this evaluation paradigm, EEGAgentBench provides a unified interactive environment for iterative, tool-assisted EEG analysis. Interaction follows a ReAct-style multi-turn protocol~\citep{DBLP:conf/iclr/YaoZYDSN023}. At the beginning of each episode, the environment provides the task description, recording metadata, and output specification. Instead of exposing raw EEG waveforms, recordings are accessed exclusively through a unified tool interface. 
At each interaction step, the agent may invoke an EEG analysis tool with user-defined arguments or terminate by submitting a final prediction. 
The environment validates each request, executes the corresponding deterministic signal-processing or statistical computation, and returns the resulting observations. 
All tools perform deterministic analysis without task-specific classifiers or label information. The benchmark therefore evaluates the agent's ability to retrieve and reason over evidence rather than rely on learned prediction modules.  Invalid requests (e.g., unavailable channels, out-of-range time intervals, or unsupported analyses) return standardized error responses.

The environment exposes 10 EEG analysis tools covering the principal operations commonly used in EEG interpretation. This tool set includes recording metadata (sampling rate, duration, channel names, and channel types), signal quality assessment (e.g., root mean square amplitude, flat-line ratio, extrema repetition ratio, power-line noise, and high-frequency noise), frequency- and time-domain analysis (e.g., power spectrum, band power, Hjorth parameters, line length, and zero-crossing rate), windowed feature extraction for temporal analysis, transient candidate detection (time, amplitude, width, and steepness of candidate waveforms), and cross-channel analysis (left--right asymmetry and inter-channel correlation).  Collectively, they provide complementary observations spanning signal characteristics, temporal dynamics, and spatial relationships. The same tool suite is shared across all signal-analysis tasks (T2--T6), allowing agents to construct task-specific workflows without predefined pipelines or task-specific interfaces.

\subsection{Benchmark Construction}

We construct 1,703 candidate instances from six existing public datasets: MedMCQA~\citep{DBLP:conf/chil/PalUS22}, EEGdenoiseNet~\citep{zhang2021eegdenoisenet}, Bonn~\cite{bonn}, OpenNeuro ds004504~\citep{DBLP:journals/data/MiltiadousTAIGTATGGT23}, CHB-MIT~\citep{shoeb2009application,goldberger2000physiobank}, and Sleep-EDFx~\citep{DBLP:journals/tbe/KempZTKO00,goldberger2000physiobank}. We convert questions, recordings, and annotations from these sources into a common format. This format includes a task description, recording information, and a structured output specification.

\begin{figure}[!t]
\centering
\includegraphics[width=0.9\linewidth]{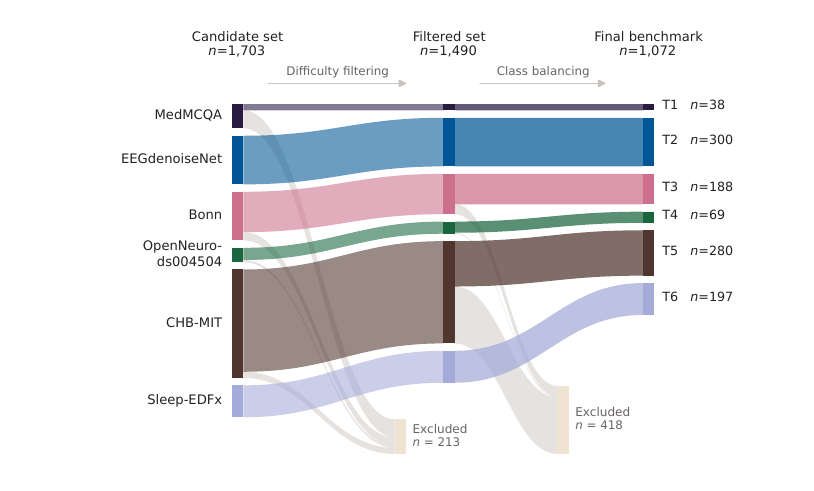}
\caption{Dataset construction and filtering process.}
\label{fig:construction}
\end{figure}

As shown in Figure~\ref{fig:construction}, benchmark construction consists of two stages. 
First, a screening panel removes overly easy instances. The panel includes models of different sizes and families. It uses instance-level completion results to identify these instances, which reduces the candidate set from 1,703 to 1,490. Second, deterministic class balancing based on task-label distributions produces the final 1,072 evaluation instances. For T6, recordings is unchanged, but we trim some continuous wake epochs at the beginning and end of selected recordings together with the corresponding signals. This reduces the extent to which long wake periods dominate full-night sleep staging. Detailed construction and filtering rules are provided in Section~\ref{app:construction}.

\FloatBarrier

\section{Experiments}
\noindent\begin{minipage}{\linewidth}
We evaluate current LLM agents on EEGAgentBench across diverse EEG analysis tasks, temporal horizons, and interaction settings. Our experiments address four questions:
\begin{enumerate}
    \item \textbf{RQ1:} How well do current LLM agents perform on EEGAgentBench?
    \item \textbf{RQ2:} Does EEGAgentBench effectively distinguish agent capabilities across diverse EEG tasks?
    \item \textbf{RQ3:} Does model scale or inference cost consistently predict performance?
    \item \textbf{RQ4:} What are the major limitations of current agents on long-horizon EEG analysis?
\end{enumerate}
\end{minipage}

\begin{figure}[!t]
\centering
\includegraphics[width=0.8\linewidth]{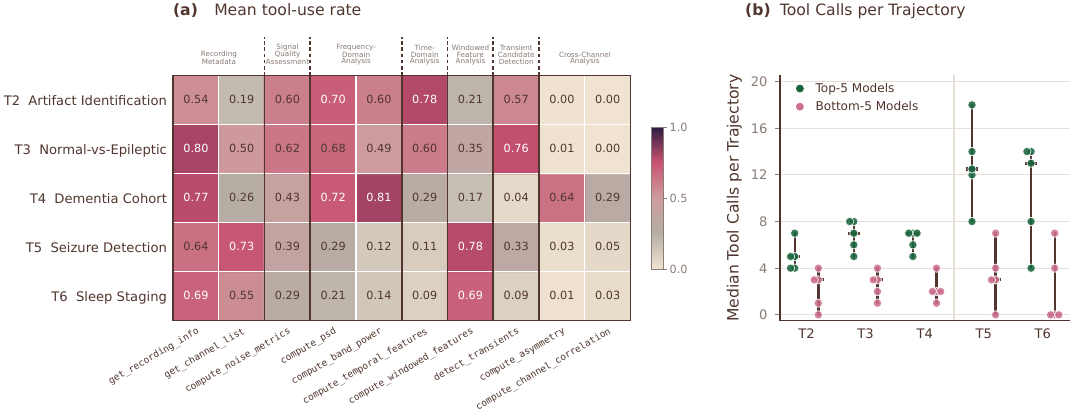}
\caption{Tool-use behavior across the five signal tasks. (a) Mean tool-use rate, averaged equally across 29 models. (b) Per-model median tool calls per trajectory for fixed Top-5 and Bottom-5 groups ranked by Overall.}
\label{fig:tool_use_behavior}
\end{figure}

\subsection{Experimental Setup}

We evaluate 29 LLMs from 15 model families on EEGAgentBench: DeepSeek-V3.2, DeepSeek-V4-Flash, and DeepSeek-V4-Pro; ERNIE-4.5-21B-A3B and ERNIE-4.5-VL-424B; Gemini-2.5-Pro and Gemini-3.5-Flash; GLM-5.1 and GLM-5.2; GPT-5.4, GPT-5.5, and GPT-5.6-sol; Grok-4.5, Hy3, Kimi-K2.6, Llama3.3-70B, LongCat-2, MiMo-V2.5-Pro, MiniMax-M3, Mistral-Medium-3.5, and Claude-Opus-4.6; and Qwen3-1.7B, Qwen3-4B, Qwen3-8B, Qwen3.5-35B-A3B, Qwen3.5-122B-A10B, Qwen3.5-397B-A17B, Qwen3.6-27B, and Qwen3.7-Max.

Each model is evaluated independently on all 1,072 instances across the six tasks. For T1, models answer EEG knowledge questions directly. For T2--T6, every model has access to the same set of 10 EEG analysis tools and may iteratively invoke them before submitting a final answer. To prevent non-convergent trajectories, we impose a maximum interaction limit. Unless otherwise specified, results are reported from a single run for each model--instance pair. Final predictions are evaluated using the task-specific metrics defined in Sections~\ref{app:metrics}--\ref{app:dice_seizure}. The Overall score is the equally weighted mean of T1 accuracy, Macro-F1 for T2--T4 and T6, and Event-F1 for T5. Tool selection, invocation order, and intermediate observations do not directly contribute to task scores, but complete interaction trajectories are analyzed to characterize tool-use behavior, inference efficiency, and failure modes. Model versions, parameter counts, settings, interaction limits, and statistical methods are provided in Section~\ref{app:model_sizes}.

\subsection{Main Results}
\label{sec:main_results}

\textbf{Finding 1: No single model dominates EEGAgentBench across all tasks}. Table~\ref{tab:main_results} summarizes the main results for all 29 models. Using the six primary metrics to compute Overall, GPT-5.6-sol ranks first with 0.631, followed by MiniMax-M3 with 0.530 and Qwen3.7-Max with 0.529. GPT-5.6-sol achieves the best primary score on T4--T6, whereas T1--T3 are led by Llama3.3-70B, Qwen3.5-397B-A17B, and Qwen3.7-Max, respectively. Although MiniMax-M3 does not rank first on any single task, its balanced performance makes it the strongest open-weight model.

\begin{figure}[H]
\centering
\includegraphics[width=0.90\textwidth]{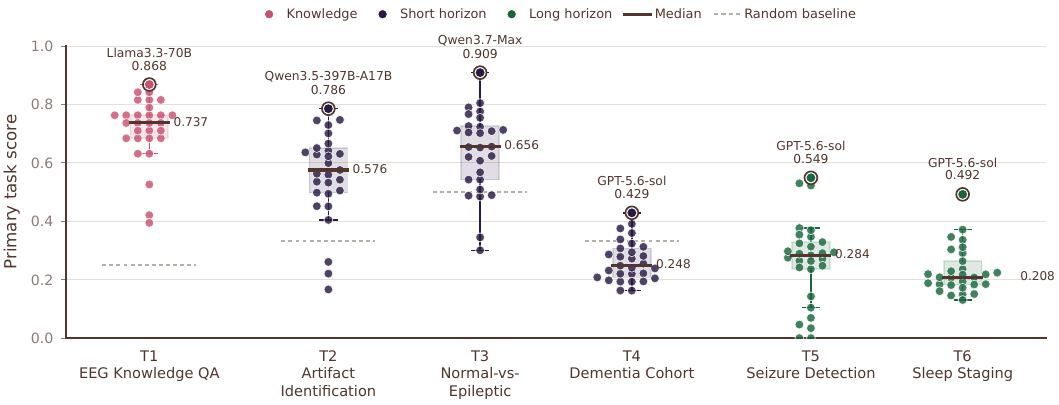}
\caption{Task-level performance distributions across 29 models using the six primary metrics that form Overall. Horizontal jitter only separates overlapping points. Boxes show medians and interquartile ranges; whiskers extend to 1.5 IQR, and labeled points mark task leaders. Dashed lines indicate uniform-random baselines for T1--T4; no random baseline is shown for T5 or T6.}
\label{fig:task_score_distributions}
\end{figure}

\textbf{Finding 2: The task distributions separate areas of model competition from shared bottlenecks. }Figure~\ref{fig:task_score_distributions} shows that the six tasks exhibit distinct performance distributions. T1 scores are generally high, suggesting that most models have acquired the benchmarked EEG knowledge. T2 and T3 have high upper bounds but wide score distributions, making them effective at distinguishing short-horizon signal analysis capability. In contrast, T4 and T6 show consistently low performance across models, indicating that dementia cohort classification and full-night sleep staging remain shared bottlenecks. T5 falls between these patterns: its median Event-F1 is only 0.284, two models fail to detect any true event, whereas the top three models all exceed 0.52.

\textbf{Finding 3: Long-horizon tasks change tool-use strategies and widen the difference in interaction depth.}  Figure~\ref{fig:tool_use_behavior}(a) shows that agents adopt distinct tool-use strategies across the five signal-analysis tasks rather than relying on a fixed workflow. T2 and T3 rely mainly on frequency-domain analysis, time-domain analysis, and transient candidate detection. T4 places more emphasis on band power and left--right asymmetry, whereas T5 and T6 shift clearly toward windowed feature analysis. These task-specific patterns suggest that agents adjust their tool choices to the analysis goal instead of mechanically repeating a fixed workflow. Figure~\ref{fig:tool_use_behavior}(b) further shows a difference in typical call depth between the Top-5 and Bottom-5 models even on short-horizon tasks; on T5 and T6 (long-horizon task), the gap widens to 12.5 versus 3 calls and 13 versus 0 calls, respectively. Leading models usually sustain longer analysis processes, whereas weaker models often end their trajectories after only a few calls.

\subsection{Model Scale and Performance}
\label{sec:model_scale}

\textbf{EEGAgentBench captures broad scaling trends without exhibiting a rigid parameter threshold.} Figure~\ref{fig:model_size_performance} shows that models below 32B generally obtain lower Overall scores, with Qwen3-1.7B, Qwen3-4B, Qwen3-8B, ERNIE-4.5-21B-A3B, and Qwen3.6-27B scoring between 0.228 and 0.433. Although the strongest model in this range, Qwen3.6-27B, achieves performance comparable to several substantially larger models, it still trails MiniMax-M3, the best-performing open-weight model, by 0.098. These results indicate that EEGAgentBench reflects the overall benefits of scaling while avoiding a trivial size-based ranking.

\textbf{EEGAgentBench distinguishes frontier models beyond parameter scale.} Among larger models, parameter count becomes less informative about performance. Llama3.3-70B reaches an Overall score of 0.429, close to the two trillion-scale models with reported parameter counts: LongCat-2 at 0.445 and DeepSeek-V4-Pro at 0.439. MiniMax-M3 is the best-performing open-weight model at 0.530, outperforming all four models with reported parameter counts of at least 1T. Within the 100B--1T range, Overall spans from 0.402 for ERNIE-4.5-VL-424B to 0.530 for MiniMax-M3.

\textbf{EEGAgentBench is sensitive to fine-grained capability differences within model families.} In the dense Qwen3 series, Overall improves from 0.233 (1.7B) to 0.299 (4B) and 0.332 (8B). Similarly, the Qwen3.5 MoE series increases from 0.459 (35B-A3B) to 0.485 (122B-A10B) and 0.501 (397B-A17B). Although performance consistently improves with scale within each family, the much larger variation observed across model families indicates that scaling alone is insufficient to explain agent capability.

\subsection{Cost--Performance Trade-offs}
\label{sec:cost_tradeoffs}

We estimate the LLM inference cost of completing EEGAgentBench evaluation from each model's input/output token usage over 1,072 instances and its public token prices. The analysis covers 19 models with comparable pricing information. All input tokens are charged at the standard input rate without cached-input discounts. The estimates exclude tool execution and hardware costs. Full pricing information and calculation procedures are provided in Section~\ref{app:pricing}.

\begin{figure}[!t]
\centering
\includegraphics[width=0.95\linewidth]{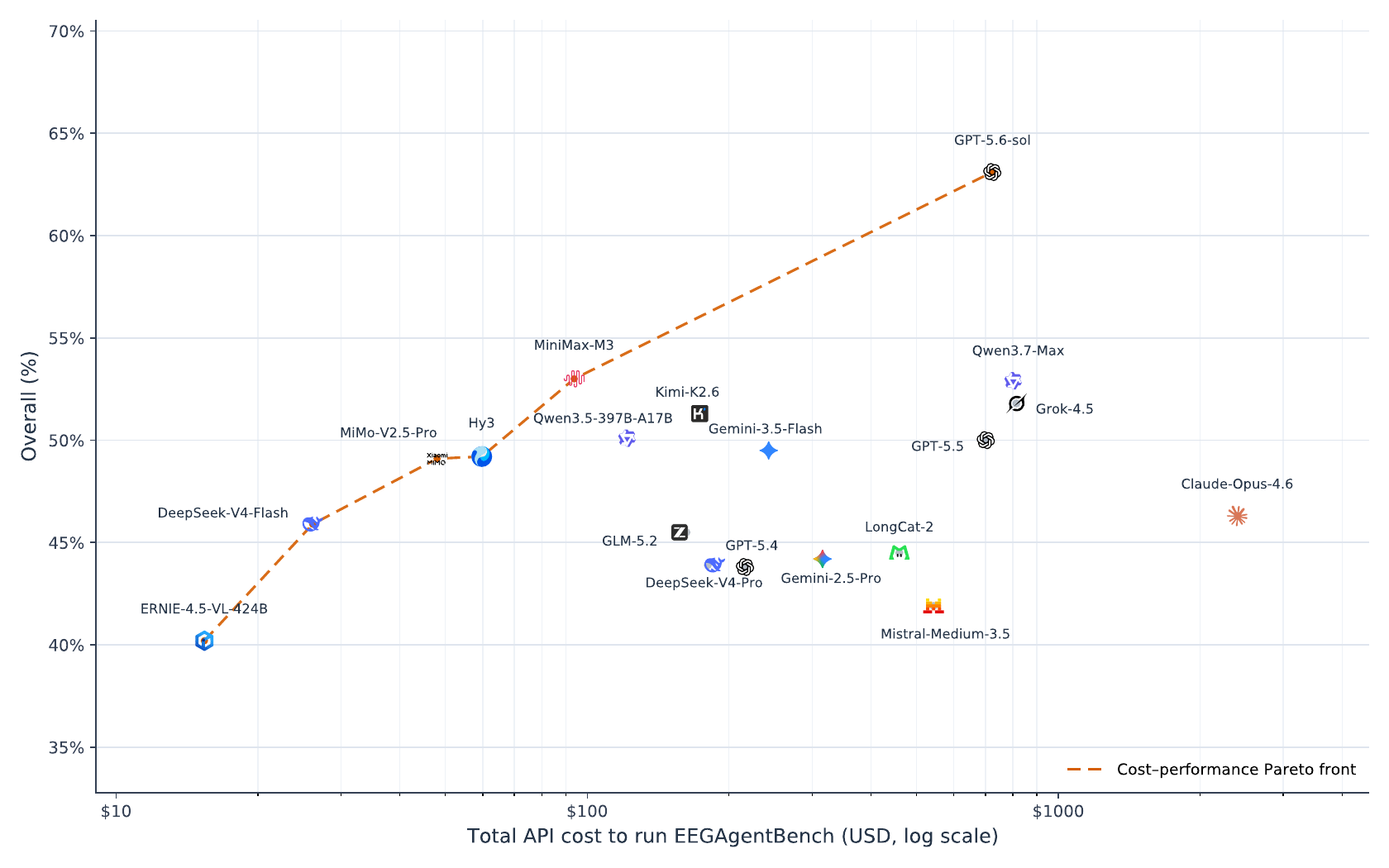}
\caption{Estimated cost versus Overall for 19 models. The dashed line marks the cost--performance Pareto frontier.}
\label{fig:cost_performance}
\end{figure}

Figure~\ref{fig:cost_performance} shows that the estimated cost of a full EEGAgentBench evaluation ranges from \$15.4 to \$2,396.5, spanning more than two orders of magnitude. ERNIE-4.5-VL-424B, DeepSeek-V4-Flash, MiMo-V2.5-Pro, MiniMax-M3, and GPT-5.6-sol form the cost--performance Pareto frontier. Along this frontier, Overall rises from 0.402 to 0.530 as cost increases from \$15.4 to \$93.8. GPT-5.6-sol further raises Overall to 0.631, but its estimated cost reaches \$723.9.


Higher cost does not consistently lead to better performance on EEGAgentBench. MiniMax-M3 reaches an Overall score of 0.530 at about 13\% of the cost of GPT-5.6-sol. Qwen3.7-Max and Grok-4.5 both cost more than \$800 but reach Overall scores of 0.529 and 0.518, respectively. Claude-Opus-4.6 has the highest estimated cost, but its Overall score is 0.463.

\subsection{Error Analysis of Long-Horizon EEG Tasks}
\label{sec:long_horizon_errors}

\begin{figure}[!t]
\centering
\includegraphics[width=0.95\linewidth]{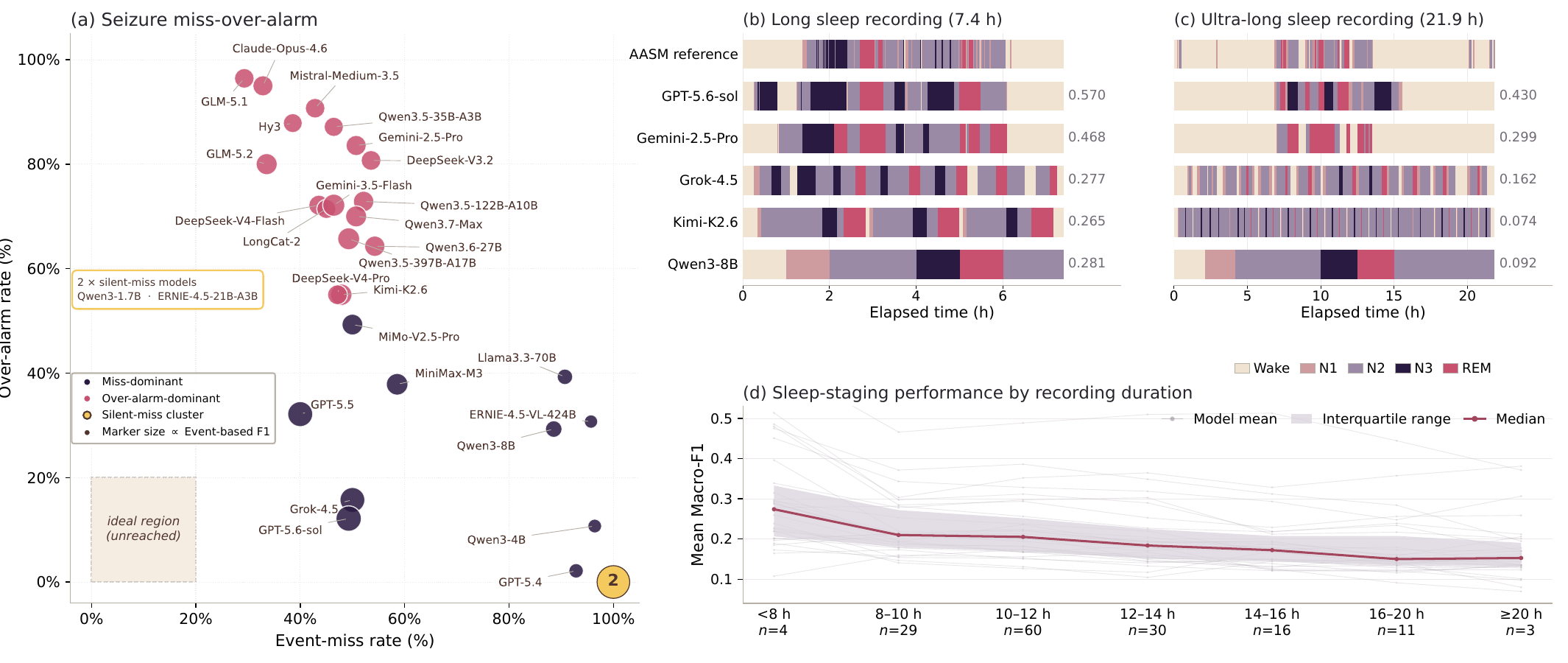}
\caption{Long-horizon EEG error analysis. (a) Event-miss rate on seizure-positive records versus over-alarm rate on seizure-free records for T5. (b--c) Hypnograms from five agents on 7.4-hour and 21.9-hour Sleep Cassette records; numbers denote recording-level Macro-F1. (d) Per-model Macro-F1 across seven recording-duration bins for T6. Thin lines denote model means, the band denotes the interquartile range across models, and dark line denotes median.}
\label{fig:long_horizon_errors}
\end{figure}

Figure~\ref{fig:long_horizon_errors}(a) compares the event-miss rate on seizure-positive recordings with the over-alarm rate on seizure-free recordings, where a miss indicates that no predicted interval overlaps a reference seizure. Most models exhibit a trade-off: lower miss rates are typically accompanied by higher false-alarm rates. GPT-5.6-sol achieves a low over-alarm rate of 12.1\% but still misses 49.3\% of seizure-positive recordings. In contrast, GLM-5.1 reduces the miss rate to 29.3\% but produces false alarms on 96.4\% of seizure-free recordings. Two models never predict seizures, and four others miss over 90\% of seizure-positive recordings. No current model achieves both low miss and low over-alarm rates.

Longer recordings remain a major challenge for sleep staging (T6). Figure~\ref{fig:long_horizon_errors}(d) summarizes case-level Macro-F1 on the Sleep Cassette subset by recording duration. From recordings shorter than 8 hours to those at least 20 hours long, performance declines for 28 of 29 models, with a median Macro-F1 drop of 0.123 (44.4\% relative). MiMo-V2.5-Pro is the only exception, improving by just 0.005. The representative cases in Figure~\ref{fig:long_horizon_errors}(b--c) show that performance deteriorates even when every epoch is assigned a label: GPT-5.6-sol drops from 0.570 to 0.430, while Kimi-K2.6 declines from 0.265 to 0.074 as recording duration increases from 7.4 to 21.9 hours. These results highlight the difficulty of maintaining reliable reasoning over ultra-long recordings, even for frontier LLMs with million-token context windows.

\section{Conclusion}

We introduced EEGAgentBench, the first unified benchmark for evaluating LLM agents on short- and long-horizon EEG analysis within a shared interactive environment. By standardizing diverse EEG tasks, tool interactions, and evaluation protocols, EEGAgentBench enables comprehensive assessment of reasoning, tool use, and long-horizon analysis capabilities. Our evaluation of 29 frontier LLMs demonstrates the discriminative power of EEGAgentBench. The benchmark distinguishes agent capabilities beyond model scale and inference cost. The results expose clear limitations in long-horizon EEG analysis. We hope EEGAgentBench provides a standardized foundation for developing and evaluating the next generation of autonomous EEG agents.

\bibliographystyle{unsrtnat}
\bibliography{main}

\clearpage
\beginappendix

\section{Appendix}

\subsection{Benchmark Construction and Filtering}
\label{app:construction}

\paragraph{Overview.} The six tasks share a single construction pipeline with three parts: (i) assembling a candidate pool from six public data sources; (ii) a two-stage instance selection that first removes overly easy instances by difficulty and then balances classes; and (iii) task-specific data preparation. The candidate pool contains 1{,}703 instances; difficulty screening reduces it to 1{,}490, and class balancing yields the final 1{,}072 evaluation instances. The dataset-construction figure in the main paper shows the per-task flow through these steps. Below we describe the screening panel, the two stages, and the task-specific preparation in turn.

\paragraph{Difficulty screening.} We estimate the difficulty of each instance with a screening panel of five models spanning different sizes and families: Qwen3-4B, Qwen3-8B, Qwen3.5-35B-A3B, DeepSeek-V3.2, and MiMo-V2.5-Pro, with parameter counts from 4B to about 1T. For each instance, we record the fraction of panel models that answer it correctly as its solve rate, and we treat an instance with a solve rate of at least 0.9 as overly easy; with five panel models, this means all five must solve the instance. A correct answer follows each task's primary scoring: a correct option for T1, a correct class label for T3 and T4, and for T5 a Dice@seizure of at least 0.30 on seizure-positive records and no false alarm on seizure-free records (Dice@seizure is defined in Section~\ref{app:dice_seizure}). This stage removes 213 overly easy instances in total: 111 from T1, 51 from T3, 13 from T4, and 38 from T5, where all 38 come from seizure-free records, indicating that instances solvable by staying silent are concentrated on the seizure-free side. T2 is a constructed balanced set and the full-night recordings of T6 are handled separately, so neither is filtered at this stage. The candidate pool is thus reduced from 1{,}703 to 1{,}490 instances.

\paragraph{Class balancing.} For classification tasks whose primary metric is accuracy or macro-F1, an imbalanced label distribution lets a model score highly by favoring the majority class, so we balance these tasks by their label distribution; the balancing is deterministic and reproduces the same evaluation set on repeated runs. For T3 we target an equal number of normal and epileptic records and an equal split between the two normal subsets (eyes-open and eyes-closed healthy recordings), giving 47 eyes-open, 47 eyes-closed (94 normal in total), and 94 seizure records, for 188 instances; when a group exceeds its target, we remove its easiest instances first. For T4 we use the smallest class, FTD, as the target count and take 23 instances per class, giving 23 each for AD, FTD, and HC, for 69 instances. For T5 we balance to a 1:1 ratio: we keep all 140 seizure-positive records and downsample the 529 seizure-free records to 140 by removing the easiest instances first, yielding 280 instances. T1 cannot be balanced because its subject distribution is highly skewed (most subjects have only one or two questions), so it passes through only the first stage and keeps 38 instances; its options are deterministically reshuffled to remove option-position bias. T2 is already a constructed balanced set and keeps 300 instances. After this stage the evaluation set is fixed at 1{,}072 instances.

\paragraph{Task-specific data preparation.} Three tasks require additional preparation. (1)~\emph{Bonn subset selection for T3.} The Bonn dataset contains five subsets of single-channel EEG segments. Because T3 is a binary normal-versus-epileptic classification, we use only the three subsets with unambiguous labels: healthy-volunteer scalp EEG (eyes-open and eyes-closed) as normal, and ictal (seizure) recordings as epileptic. The other two subsets are interictal recordings: the patient has epilepsy, but the segment contains no seizure; their binary label is therefore ambiguous. We exclude them to keep the two classes clean. This restricts the epileptic class to ictal segments, making the task closer to healthy versus actively seizing. (2)~\emph{Semi-synthetic construction for T2.} T2 contains 100 clean, 100 ocular, and 100 muscle segments (300 in total); each contaminated class spans ten signal-to-noise ratio levels from $-7$ to $+2$~dB with 10 segments per level. The mixing formula and the definition of signal-to-noise ratio are given in Section~\ref{app:t2_snr}. (3)~\emph{Edge-wake trimming for T6.} Sleep staging is highly class-imbalanced: wake accounts for 63.7\% of the labels before trimming (a global proportion of 0.637), and home-based full-night recordings contain long wake periods before sleep onset and after final awakening that let wake dominate macro-F1. We therefore trim the leading and trailing runs of wake from each recording under four rules. First, we trim data and labels together: labels are sliced by the corresponding interval, and the signal is losslessly cut from the source EDF along its 30-second data-record boundaries, with the recording then pointing to the trimmed file, from which evaluation reads the trimmed duration and signal. Second, we trim only the leading and trailing wake; if the first or last epoch at one end already belongs to one of the other four stages, that end is not trimmed. Third, we trim only the contiguous wake runs at the two ends and keep any wake inside the recording. Fourth, we set the target so that the post-trim wake proportion approaches that of the largest of the other four stages. Under these rules the 197 recordings fall into three cases:
\begin{itemize}
    \item 42 recordings already meet the target and are not trimmed;
    \item 115 reach the target by trimming leading and trailing wake;
    \item 40 whose interior wake is too large to reach the target even after fully trimming the ends retain a small wake buffer next to the sleep body at each end.
\end{itemize}
Because cuts fall on data-record boundaries, almost all recordings align naturally with 30-second epochs. For the single recording with 60-second data records, we align the cut to an even number of epochs; this avoids cutting into sleep at the cost of slightly less trimming. In the end 155 recordings are trimmed, the global epoch count drops from 461{,}702 to 264{,}422 (a 42.7\% reduction), and the wake proportion drops from 0.637 to 0.365, closer to balance across the four remaining stages.

\subsection{Evaluation Metrics}
\label{app:metrics}

\paragraph{Overview.} Every task is scored automatically from the agent's final structured output. A missing or unparseable label is counted as incorrect for T1--T4. The two structured long-horizon tasks use task-specific fallbacks: an empty T5 event list is a valid no-seizure prediction, while a missing, empty, or unusable T6 hypnogram is expanded to an all-wake sequence and scored normally. The metrics used across the six tasks are as follows: T1 uses accuracy; T2--T4 use accuracy and macro-F1; T5 uses Event-F1 and Dice@seizure; and T6 uses macro-F1 and Cohen's kappa. This subsection defines accuracy, macro-F1, and Cohen's kappa, which are shared across several tasks; the two T5 metrics require separate definitions in Sections~\ref{app:event_f1} and~\ref{app:dice_seizure}.

\paragraph{Accuracy.} Accuracy is the mean of a per-instance 0/1 correctness over the scored instances,
\begin{equation}
\text{Accuracy}=\frac{1}{N}\sum_{i=1}^{N}\mathbb{1}[\hat{y}_i=y_i],
\end{equation}
where $y_i$ is the reference label, $\hat{y}_i$ is the model prediction, and $N$ is the number of scored instances. For T1, we extract the chosen option letter (A--D) from the reply---accepting forms such as ``B'', ``(B)'', ``Option B'', or a verbatim restatement of the option text---and match it exactly against the reference option; a reply from which no option can be recovered, or an empty reply, is counted as incorrect. For T2--T4, we match the predicted class label against the reference label after lower-casing and stripping whitespace (clean/ocular/muscle for T2, normal/epileptic for T3, and AD/FTD/HC for T4). The uniform-random accuracy baselines are 0.25 for T1 (four options), $1/3$ for the balanced three-class T2 and T4, and 0.5 for the balanced two-class T3.

\paragraph{Macro-F1.} Macro-F1 is computed from the confusion matrix over the scored subset. For each class $c$, with true positives $TP_c$, false positives $FP_c$, and false negatives $FN_c$, the per-class F1 is
\begin{equation}
F1_c=\frac{2\,TP_c}{2\,TP_c+FP_c+FN_c},
\end{equation}
and macro-F1 is their unweighted mean $\frac{1}{|C|}\sum_{c\in C}F1_c$, where $C$ is the set of reference labels for the task. For T2--T4, the classes are the task's label set; for T6, the classes are the five sleep stages scored per epoch (see below). Unlike accuracy, macro-F1 weights every class equally and therefore penalizes single-class strategies: on the balanced T4, for example, always predicting AD gives an accuracy of about $1/3$ but a macro-F1 of only 0.167.

\paragraph{Cohen's kappa.} T6 outputs a per-epoch sequence of sleep stages (one label per 30-second epoch), and we use per-epoch Cohen's kappa to measure staging agreement after discounting chance agreement,
\begin{equation}
\kappa=\frac{p_o-p_e}{1-p_e},
\end{equation}
where $p_o$ is the observed per-epoch agreement (equal to the per-epoch accuracy) and $p_e=\sum_{c}\big(\tfrac{n^{\text{pred}}_c}{N}\big)\big(\tfrac{n^{\text{gold}}_c}{N}\big)$ is the chance agreement computed from the marginal class frequencies of the prediction and the reference, with $c$ ranging over the five sleep stages. $\kappa$ can be negative when the prediction departs systematically from the reference; in the degenerate case where a single class fills both the prediction and the reference so that $p_e=1$, we score a perfect match as 1 and otherwise as 0. Because wake accounts for a large share of the labels (Section~\ref{app:construction}), macro-F1 alone can be inflated by correctly labeling wake epochs, whereas kappa removes this chance baseline and thus exposes near-random staging.

\paragraph{Hypnogram alignment.} The T6 prediction is given as a run-length encoding, a list of segments each with a start epoch, an end epoch, and a stage. Before computing macro-F1 and kappa, we expand it into a per-epoch label sequence, handling common issues in real LLM outputs tolerantly: gaps are filled with wake, overlapping or out-of-order segments are resolved by keeping the last-written label, out-of-range or unrecognized stages are mapped to wake, malformed rows are skipped, any tail beyond the total epoch count is truncated, and an empty encoding is treated as an all-wake sequence. The predicted sequence is then aligned to the reference length---truncated if longer and padded with wake if shorter---and the aligned per-epoch sequence is used in the two equations above.

\subsection{Event-F1 for T5 Seizure Detection}
\label{app:event_f1}

For T5 seizure event detection, we score each model with a SzCORE-style event-based F1, following the seizure event scoring of the SzCORE evaluation framework and its \tool{timescoring} reference implementation~\citep{DBLP:journals/corr/abs-2402-13005}. This metric measures whether each clinical seizure event is detected and how many false alarms a model raises. It is computed from the reference seizure intervals and the intervals predicted by the agent. We use an independent continuous-interval implementation that quantizes interval boundaries to 0.1~s and follows the SzCORE parameters and processing order; we therefore call the metric SzCORE-style. Table~\ref{tab:event_f1_params} lists the default parameters.

\begin{table}[!t]
\caption{Default parameters of the SzCORE-style event-based F1 used for T5.}
\label{tab:event_f1_params}
\centering
\small
\begin{tabular}{@{}p{0.24\columnwidth} p{0.13\columnwidth} p{0.49\columnwidth}@{}}
\toprule
Parameter & Value & Role \\
\midrule
Minimum overlap & 0 s & A prediction matches a reference event if they overlap for any positive duration \\
Tolerance before start & 30 s & Extend each reference event 30 s before its start \\
Tolerance after end & 60 s & Extend each reference event 60 s after its end \\
Minimum gap between events & 90 s & Merge events whose gap is strictly shorter than 90 s \\
Maximum event duration & 300 s & Split events longer than 300 s into consecutive events \\
Time resolution & 10 Hz & Quantize interval boundaries to 0.1 s \\
\bottomrule
\end{tabular}
\end{table}

For a single record, the scoring runs in the following order. The matching between reference and predicted events is not one-to-one: one prediction may detect several reference events, and several predictions inside the tolerance window of one reference event do not add extra false positives. (1)~Parse the reference intervals and the predicted intervals, clip their start and end times to the record range, and quantize them to 0.1~s. (2)~Merge events separately within the reference set and within the prediction set: two neighboring events are merged when the gap between them is strictly shorter than 90~s, and a gap of exactly 90~s is not merged. (3)~Split any reference or predicted event longer than 300~s into consecutive shorter events. (4)~Extend each processed reference event 30~s before its start and 60~s after its end, and clip the result to the record range. (5)~For each extended reference event, check whether any predicted event overlaps it for a positive duration; count one true positive (TP) if so, and one false negative (FN) otherwise. (6)~For each processed predicted event, count one false positive (FP) if it does not overlap any extended reference event.

The event-based scoring does not define true negatives. From the accumulated event counts, the metrics are
\begin{equation}
\mathrm{Sensitivity}=\frac{TP}{TP+FN},\qquad
\mathrm{Precision}=\frac{TP}{TP+FP},
\end{equation}
\begin{gather}
\text{Event-F1}=\frac{2\,TP}{2\,TP+FP+FN},\\
\mathrm{FP/24h}=\frac{FP}{\text{duration in hours}/24}.
\end{gather}

CHB-MIT contains several records for the same subject. To keep subjects with more records from carrying more weight, we aggregate by subject. For each subject, we sum the TP, FP, FN, and recording duration over all of that subject's records, and compute Event-F1, Sensitivity, Precision, and FP/24h from these subject-level counts. The reported Event-F1 is the unweighted mean over the 24 subjects. Because this scoring does not define true negatives, a model that never detects any reference event receives an Event-F1 of 0, so almost never predicting a seizure is not rewarded.

\setcounter{dbltopnumber}{2}
\renewcommand{\dbltopfraction}{0.95}
\renewcommand{\dblfloatpagefraction}{0.85}
\renewcommand{\textfraction}{0.05}

\subsection{Dice@seizure for T5 Seizure Detection}
\label{app:dice_seizure}

Event-F1 measures whether each reference seizure event is detected under a wide tolerance window, but it does not measure how well the predicted time span matches the true one. To measure this, we report Dice@seizure (Dice-S in the main-results table), which quantifies the temporal overlap between the predicted seizure spans and the reference seizure spans along the time axis. It is computed from the same reference and predicted intervals as Event-F1, but it does not use the tolerance windows, boundary extension, or event merging and splitting of the SzCORE-style scoring. It compares the two sets of intervals directly in seconds.

We first define a per-record score. For one record, let $R$ be the set of reference seizure intervals and $P$ the set of predicted intervals, each interval given as an onset and offset in seconds. Within each set we merge intervals that overlap or touch into a disjoint union, so that overlapping predictions are not counted twice. Let $|R|$ and $|P|$ denote the total number of seconds covered by the merged reference set and the merged prediction set, and let $|R\cap P|$ denote the number of seconds covered by both. The per-record temporal Dice is
\begin{equation}
\mathrm{Dice}=\frac{2\,|R\cap P|}{|R|+|P|}.
\end{equation}
This is the Dice overlap of two binary masks on the time axis, where each mask marks the seconds labeled as seizure. It equals $1$ when the predicted span exactly matches the reference span and $0$ when they do not overlap at all. We use two conventions for the boundary cases. When both $R$ and $P$ are empty, the record is scored as $1$, because the agent correctly reported no seizure. When $R$ is non-empty but $P$ is empty, the record is scored as $0$, because the agent missed the seizure entirely. The score also penalizes over-wide predictions: when a single interval covers the whole record, $|P|$ increases while $|R\cap P|$ remains equal to $|R|$, so the Dice value approaches $0$.

Because a reference span is needed for the overlap to be meaningful, Dice@seizure is computed only on the seizure-positive records, that is, the $140$ T5 records whose reference contains at least one seizure interval. On seizure-free records the reference is empty and the per-record Dice degenerates to $1$ when the agent stays silent and $0$ when it predicts anything, so false-alarm behavior is not summarized through Dice. False predicted events contribute to Event-F1 through its FP count and to the duration-normalized FP/24h defined above. The main-paper error analysis additionally reports the over-alarm rate, defined as the fraction of seizure-free records with at least one predicted interval; this is equal to one minus record-level specificity. Let $S$ be the set of seizure-positive records. The reported metric is the unweighted mean of the per-record Dice over this subset,
\begin{equation}
\text{Dice@seizure}=\frac{1}{|S|}\sum_{r\in S}\mathrm{Dice}_r,
\end{equation}
where $\mathrm{Dice}_r$ is the per-record temporal Dice of record $r$. This average is taken over records rather than over subjects, so each seizure-positive record contributes equally.

\subsection{Model Parameter Counts}
\label{app:model_sizes}

Table~\ref{tab:model_sizes} lists the parameter counts used for the model-scale analysis and corresponding figure in the main paper. Counts are taken from public model information. For dense models we report the total parameter count, and for mixture-of-experts (MoE) models whose activation is reported we use the total-B--activated-B format (for example, 397B-A17B), where the number after A is the activated parameters per inference. MoE models whose activation is not reported show only the total count. Among models with reported sizes, the range spans from 1.7B to 1{,}600B (1.6T). Eight proprietary models do not have publicly reported parameter counts and are marked as not reported; these models appear in the unknown-scale band of the main-paper figure. Table~\ref{tab:model_sizes} also lists each model's context window, taken from public model information. The context window ranges from 32K tokens for the small Qwen3 models to 1M tokens for many recent large models, and we use each model's own context window during evaluation. This matters for the long-horizon tasks T5 and T6, where recordings and accumulated tool results can produce inputs of tens of thousands of tokens or more.

\begin{table}[!t]
\caption{Parameter counts and context windows of the 29 evaluated models.}
\label{tab:model_sizes}
\centering
\small
\begin{tabular*}{\linewidth}{@{\extracolsep{\fill}}l l r r@{}}
\toprule
Model & Family & Parameters & Context \\
\midrule
Qwen3-1.7B & Qwen & 1.7B & 32K \\
Qwen3-4B & Qwen & 4B & 32K \\
Qwen3-8B & Qwen & 8B & 32K \\
ERNIE-4.5-21B-A3B & ERNIE & 21B-A3B & 128K \\
Qwen3.6-27B & Qwen & 27B & 256K \\
Qwen3.5-35B-A3B & Qwen & 35B-A3B & 256K \\
Llama3.3-70B & Llama & 70B & 128K \\
Qwen3.5-122B-A10B & Qwen & 122B-A10B & 256K \\
Mistral-Medium-3.5 & Mistral & 128B & 256K \\
DeepSeek-V4-Flash & DeepSeek & 284B-A13B & 1M \\
Hy3 & Hunyuan & 295B-A21B & 256K \\
Qwen3.5-397B-A17B & Qwen & 397B-A17B & 256K \\
ERNIE-4.5-VL-424B & ERNIE & 424B-A47B & 128K \\
MiniMax-M3 & MiniMax & 428B-A23B & 1M \\
DeepSeek-V3.2 & DeepSeek & 671B-A37B & 128K \\
GLM-5.1 & GLM & 744B-A40B & 200K \\
GLM-5.2 & GLM & 753B-A40B & 1M \\
Kimi-K2.6 & Kimi & 1{,}000B-A32B & 256K \\
MiMo-V2.5-Pro & MiMo & 1{,}020B-A42B & 1M \\
DeepSeek-V4-Pro & DeepSeek & 1{,}600B-A49B & 1M \\
LongCat-2 & LongCat & 1{,}600B-A48B & 1M \\
\addlinespace[2pt]
Gemini-2.5-Pro & Gemini & Not reported & 1M \\
Gemini-3.5-Flash & Gemini & Not reported & 1M \\
GPT-5.4 & OpenAI & Not reported & 1M \\
GPT-5.5 & OpenAI & Not reported & 1M \\
GPT-5.6-sol & OpenAI & Not reported & 1M \\
Grok-4.5 & Grok & Not reported & 500K \\
Qwen3.7-Max & Qwen & Not reported & 1M \\
Claude-Opus-4.6 & Claude & Not reported & 1M \\
\bottomrule
\end{tabular*}
\end{table}

\paragraph{Execution settings.} The model identifiers and context windows used in evaluation are listed in Table~\ref{tab:model_sizes}. Decoding followed the model-specific configurations rather than a single shared temperature or sampling setting. All tasks used a limit of 1{,}000 agent turns; T1 was answered directly without tools, while the five signal tasks used a separate ceiling of 999 tool calls. These ceilings served only as safeguards against non-convergent runs. Each model--instance pair was evaluated once, and the reported analyses are descriptive; no hypothesis tests were performed.

\subsection{Token Usage of the Full Evaluation}
\label{app:token_usage}

Table~\ref{tab:token_usage} reports the token usage of completing the full EEGAgentBench evaluation, summed over all 1,072 instances across the six tasks. For each model, we report the prompt tokens (input tokens), the completion tokens (output tokens), and their sum as all tokens, and we sort the models in descending order of all tokens. These counts are the basis of the cost estimate in the main paper. Columns are rounded independently, so prompt plus completion may not sum exactly to the displayed total.

Total token usage spans a wide range, from 7.4M for Qwen3-1.7B to 598.7M for LongCat-2, a difference of about 80 times. For almost every model, prompt tokens make up most of the total, because each tool result is fed back into the prompt over many turns of interaction, so the accumulated context grows as an agent calls more tools. Completion tokens are a small share of the total for most models; the largest completion share is Mistral-Medium-3.5 with 31.2M output tokens, and models such as Gemini-3.5-Flash, GPT-5.4, and ERNIE-4.5-21B-A3B produce about 1M output tokens. Because prompt tokens dominate, total token use is driven mainly by accumulated context, not final-answer length.

\begin{table}[!t]
\caption{Token usage for the full evaluation, in millions of tokens.}
\label{tab:token_usage}
\centering
\small
\begin{tabular*}{\linewidth}{@{\extracolsep{\fill}}l r r r@{}}
\toprule
Model & Prompt tokens (M) & Completion tokens (M) & All tokens (M) \\
\midrule
LongCat-2 & 589.8 & 8.9 & 598.7 \\
DeepSeek-V4-Pro & 405.3 & 4.5 & 409.8 \\
Claude-Opus-4.6 & 390.6 & 17.7 & 408.3 \\
GLM-5.2 & 381.4 & 26.2 & 407.6 \\
Grok-4.5 & 389.9 & 5.8 & 395.7 \\
Hy3 & 320.0 & 25.7 & 345.7 \\
DeepSeek-V4-Flash & 315.9 & 11.7 & 327.6 \\
Qwen3.7-Max & 285.1 & 11.9 & 297.0 \\
MiniMax-M3 & 273.0 & 9.9 & 282.9 \\
Qwen3.5-397B-A17B & 269.1 & 7.3 & 276.4 \\
Mistral-Medium-3.5 & 206.0 & 31.2 & 237.2 \\
Gemini-2.5-Pro & 232.4 & 2.5 & 235.0 \\
GLM-5.1 & 186.2 & 10.2 & 196.4 \\
Kimi-K2.6 & 178.9 & 16.2 & 195.0 \\
Gemini-3.5-Flash & 156.0 & 1.0 & 157.0 \\
GPT-5.6-sol & 129.7 & 2.5 & 132.2 \\
GPT-5.5 & 122.7 & 2.9 & 125.6 \\
DeepSeek-V3.2 & 97.2 & 10.7 & 107.8 \\
MiMo-V2.5-Pro & 94.3 & 8.0 & 102.3 \\
GPT-5.4 & 79.3 & 1.2 & 80.5 \\
ERNIE-4.5-21B-A3B & 74.9 & 1.0 & 75.8 \\
Llama3.3-70B & 35.6 & 2.1 & 37.7 \\
Qwen3.6-27B & 31.1 & 3.7 & 34.8 \\
ERNIE-4.5-VL-424B & 30.7 & 2.0 & 32.7 \\
Qwen3.5-122B-A10B & 24.9 & 2.4 & 27.4 \\
Qwen3.5-35B-A3B & 22.0 & 2.1 & 24.1 \\
Qwen3-8B & 12.1 & 3.2 & 15.3 \\
Qwen3-4B & 10.2 & 3.0 & 13.1 \\
Qwen3-1.7B & 5.7 & 1.7 & 7.4 \\
\bottomrule
\end{tabular*}
\end{table}

\subsection{Model Pricing for the Cost Estimate}
\label{app:pricing}

The cost estimate in the main paper combines each model's token usage with its public per-token prices. For each model, the estimated cost of the full evaluation is
\begin{equation}
\text{cost (USD)} = \frac{\text{prompt tokens}}{10^6}\cdot p_\text{in} + \frac{\text{completion tokens}}{10^6}\cdot p_\text{out},
\end{equation}
where $p_\text{in}$ and $p_\text{out}$ are the input and output prices per one million tokens. We charge all input tokens at the standard input price and do not use cached-input rates, and the estimate excludes local EEG tool execution and hardware costs. Table~\ref{tab:pricing} lists the prices used for the 19 models that have comparable public pricing, all in USD per one million tokens. LongCat-2 is converted from CNY at 6.8036~CNY per USD, the rate recorded on 2026-07-09, giving about \$0.735 input and \$2.939 output per one million tokens.

\begin{table}[!htbp]
\caption{Public input and output prices used for cost estimation, in USD per million tokens.}
\label{tab:pricing}
\centering
\small
\begin{tabular*}{\linewidth}{@{\extracolsep{\fill}}l r r@{}}
\toprule
Model & Input price & Output price \\
\midrule
GPT-5.5 & 5.0 & 30.0 \\
GPT-5.6-sol & 5.0 & 30.0 \\
Claude-Opus-4.6 & 5.0 & 25.0 \\
GPT-5.4 & 2.5 & 15.0 \\
Qwen3.7-Max & 2.5 & 7.5 \\
Grok-4.5 & 2.0 & 6.0 \\
Gemini-3.5-Flash & 1.5 & 9.0 \\
Mistral-Medium-3.5 & 1.5 & 7.5 \\
Gemini-2.5-Pro & 1.25 & 10.0 \\
Kimi-K2.6 & 0.66 & 3.41 \\
DeepSeek-V4-Pro & 0.435 & 0.87 \\
MiMo-V2.5-Pro & 0.435 & 0.87 \\
ERNIE-4.5-VL-424B & 0.42 & 1.25 \\
Qwen3.5-397B-A17B & 0.385 & 2.45 \\
GLM-5.2 & 0.35 & 1.10 \\
MiniMax-M3 & 0.30 & 1.20 \\
Hy3 & 0.14 & 0.58 \\
DeepSeek-V4-Flash & 0.077 & 0.154 \\
LongCat-2 & 0.735 & 2.939 \\
\bottomrule
\end{tabular*}
\end{table}

\subsection{Task-Level Performance Distributions}
\label{app:task_distributions}

Figure~\ref{fig:task_score_distributions} gives a cross-sectional view of the 29 model scores on each task using the same six primary metrics that make up Overall: Accuracy for T1, Macro-F1 for T2--T4 and T6, and subject-averaged SzCORE-style Event-F1 for T5. Each point denotes one model, and its horizontal displacement is used only to separate overlapping observations. Boxes span the interquartile range, thick center lines and adjacent values show the median, and whiskers extend to the most extreme observations within 1.5 times the interquartile range. White-ringed, labeled points mark the best model on each task. Dashed segments show uniform-random baselines for T1--T4. We do not show random baselines for T5 or T6 because random event intervals and sleep-stage sequences require additional assumptions about event counts, durations, class priors, or temporal dependence.

The distributions reveal distinct capability regimes; no single difficulty ordering captures them. T1 is comparatively mature, with a median of 0.737 and a best score of 0.868. T2 and T3 have medians of 0.576 and 0.656, with best observed scores of 0.786 and 0.909, respectively, but their broad distributions show that these short-horizon capabilities are not shared uniformly across models. T4 remains difficult: its median Macro-F1 is 0.248, below the balanced three-class random baseline of 0.333, and even the best model reaches only 0.429. T6 is another clear bottleneck, with a median of 0.208 and a best score of 0.492, while most models remain concentrated in the low-score region.

T5 lies between these patterns but remains limited and heterogeneous: its median Event-F1 is 0.284 and its best score is 0.549. Event-F1 does not define true negatives and therefore does not reward silence on seizure-free records, but it also does not measure the temporal precision of predicted seizure intervals; Dice@seizure and the miss--over-alarm analysis are needed to characterize those behaviors. Taken together, the differences in medians, upper bounds, and interquartile ranges show why a single Overall ranking cannot fully describe current EEG-agent capabilities.

\subsection{SNR Dose--Response Analysis on T2 Artifact Identification}
\label{app:t2_snr}

To understand how artifact strength affects T2 identification, we stratify the 300 instances by artifact type and signal-to-noise ratio (SNR). The evaluation set contains 100 clean, 100 ocular-artifact, and 100 muscle-artifact epochs. Each contaminated class contains 10 epochs at each of 10 SNR levels from $+2$ to $-7$~dB. Following the EEGdenoiseNet convention, each contaminated epoch and its scaling factor are
\begin{gather}
x_\text{noisy} = x_\text{eeg} + \lambda\, x_\text{artifact},\\
\lambda = \frac{\mathrm{RMS}(x_\text{eeg})}{\mathrm{RMS}(x_\text{artifact})\cdot 10^{\mathrm{SNR_{dB}}/10}},
\end{gather}
where $\mathrm{RMS}(x)=\sqrt{\tfrac{1}{N}\sum_i x_i^2}$ is the root mean square amplitude. Substituting $\lambda$ back, the added artifact satisfies
\begin{equation}
\mathrm{RMS}(\lambda\, x_\text{artifact}) = \mathrm{RMS}(x_\text{eeg})\cdot 10^{-\mathrm{SNR_{dB}}/10},
\end{equation}
so the SNR is equivalently
\begin{equation}
\mathrm{SNR_{dB}} = 10\log_{10}\frac{\mathrm{RMS}(x_\text{eeg})}{\mathrm{RMS}(\lambda\, x_\text{artifact})}.
\end{equation}
A larger SNR therefore means that the EEG is stronger relative to the added artifact. At $+2$~dB the artifact-to-EEG RMS ratio is approximately $0.63$, at $0$~dB the two RMS values are equal, and at $-7$~dB the ratio is approximately $5.01$. We arrange the horizontal axis from $+2$ to $-7$~dB so that artifact strength, visibility, and expected identification accuracy increase from left to right. Because each artifact--SNR cell contains 10 epochs, per-cell accuracy changes in increments of 0.1.

\begin{figure}[H]
\centering
\includegraphics[width=0.90\textwidth]{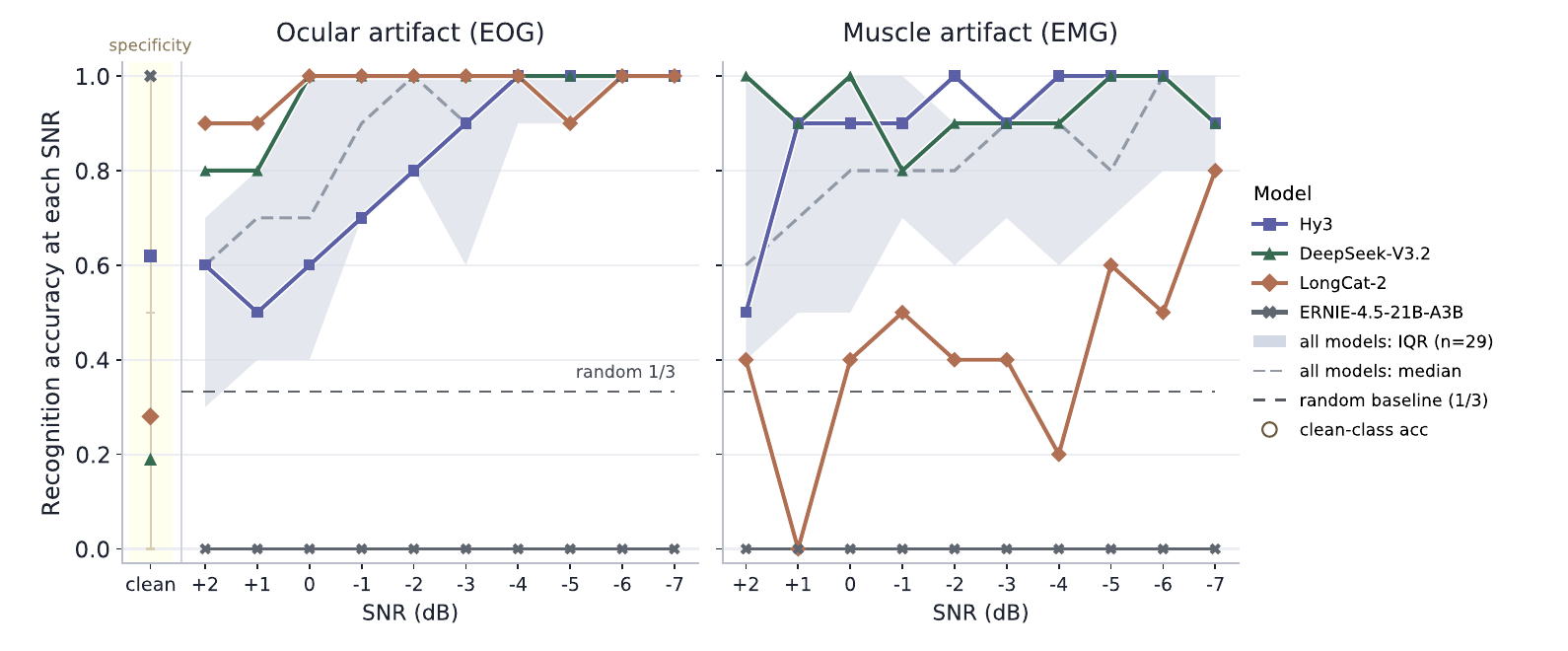}
\caption{SNR dose--response on T2 artifact identification. Left: ocular (EOG) artifacts. Right: muscle (EMG) artifacts. SNR decreases from $+2$ to $-7$~dB so that the artifact grows stronger from left to right; the vertical axis is per-class accuracy at each SNR level. Four models illustrate balanced identification, over-calling, artifact-type confusion, and failure to detect artifacts. The band and center line show the interquartile range and median across all 29 models, and the dashed line marks the uniform-random level of $1/3$. The reference column on the far left shows clean-class accuracy for the four highlighted models.}
\label{fig:t2_snr_dose_response}
\end{figure}

Figure~\ref{fig:t2_snr_dose_response} shows per-class accuracy for ocular artifacts in the left panel and muscle artifacts in the right panel. Four models are highlighted to isolate distinct error mechanisms; the light band and center line summarize the interquartile range and median across all 29 models. The dashed line marks the $1/3$ accuracy expected from uniform guessing over clean, ocular, and muscle. Clean epochs have no SNR, so the small reference column to the far left shows each highlighted model's clean-class accuracy once rather than incorporating it into either dose--response curve. This value measures how often a model correctly leaves a clean signal labeled as clean and therefore exposes over-calling of artifacts.

The results show a clear dose--response among artifact-detecting models: accuracy generally rises as SNR decreases and the artifact becomes stronger. The most discriminative region is the near-clean end from $+2$ to $0$~dB, whereas many strong models approach saturation once SNR falls below about $-3$~dB. Nevertheless, the full confusion matrices show that model behavior cannot be described by a single ability axis. It separates into one detection gate and two approximately orthogonal error axes. Two models have an artifact-detection rate below 0.5; ERNIE-4.5-21B-A3B has a detection rate of zero, predicting clean for every epoch. Among models that pass this gate, clean-class accuracy ranges from 0.77 to 0.05, exposing large differences in over-calling, while ocular--muscle type-confusion rates range from near zero to 0.54.

The four highlighted models are chosen for behavioral contrast, not ranking. Hy3 is the balanced reference, with detection rate 0.91, clean-class accuracy 0.62, and type confusion 0.07. DeepSeek-V3.2 has almost the same low type-confusion rate (0.07) and detects nearly every contaminated epoch (0.99), but its clean-class accuracy is only 0.19, identifying over-calling as the main failure. LongCat-2 also detects artifacts reliably (0.96), yet labels 54\% of muscle artifacts as ocular; this appears as high ocular accuracy but a pronounced collapse in the muscle panel. ERNIE-4.5-21B-A3B has detection rate 0 and clean-class accuracy 1, illustrating failure to cross the detection gate. Contaminated-class accuracy must therefore be read alongside clean specificity (over-calling) and cross-panel asymmetry (type confusion).

\subsection{Task Case Studies}
\label{app:cases}

\paragraph{Input templates.} To make the tool-based analysis concrete without exposing instance identifiers or record-level metadata, we show redacted input templates for all six tasks below. Placeholders in angle brackets mark values withheld from these templates. Each box corresponds to one task in the order T1 through T6.

\begin{lstlisting}[style=inboxcompact]
== T1 EEG Knowledge QA ==

## Question
Absence seizures are characterized on EEG by-

## Options
A. Hypsarrythmia
B. 1-2 Hz spike & wave
C. 3 Hz spike & wave
D. Generalized polyspikes

Reply with the single correct option letter as JSON: json
{"action": "final_answer", "answer": "<A|B|C|D>"}
\end{lstlisting}
\begin{lstlisting}[style=inboxcompact]
== T2 Artifact Identification ==

## Clinical Context
A short single-channel EEG epoch (~2s at 256 Hz). Assess its signal quality.

## EEG Record
- **Record ID**: <record_id>
- **Sampling Rate**: 256.0
- **Channels**: 1
- **Channel Names**: ['EEG']
- **N Samples**: 512
- **Duration Sec**: 2.0

## Question
Is this EEG epoch clean, contaminated by ocular (EOG) artifact, or contaminated by muscle (EMG) artifact? Use the measurement tools, then give your decision.
\end{lstlisting}
\begin{lstlisting}[style=inboxcompact]
== T3 Normal-vs-Epileptic ==

## Clinical Context
single-channel EEG recording, sampling rate 173.61 Hz, duration approximately 23.6 seconds, 4096 samples.

## EEG Record
- **Record ID**: <record_id>
- **Channels**: 1
- **Sampling Rate**: 173.61
- **N Samples**: 4096
- **Duration Sec**: 23.59

## Question
Classify this EEG recording as normal or epileptic and list the supporting findings.
\end{lstlisting}
\begin{lstlisting}[style=inboxcompact]
== T4 Dementia Cohort ==

## Clinical Context
64-year-old male participant, eyes-closed resting-state EEG recording, 19-channel 10-20 system, sampling rate 500 Hz, recording duration approximately 796 seconds.

## EEG Record
- **Record ID**: <record_id>
- **Channels**: 19
- **Channel Names**: ['Fp1', 'Fp2', 'F3', 'F4', 'C3', 'C4', 'P3', 'P4', 'O1', 'O2', 'F7', 'F8', 'T3', 'T4', 'T5', 'T6', 'Fz', 'Cz', 'Pz']
- **Sampling Rate**: 500
- **Duration Sec**: <duration_sec>
- **Montage**: 10-20
- **Reference**: A1 A2
- **Gender**: <gender>
- **Age**: <age>

## Question
Based on this eyes-closed resting-state EEG and the research context, identify the participant's cohort: Alzheimer's disease (AD), frontotemporal dementia (FTD), or healthy control (HC).  This is a cohort label, not a diagnosis.  List the supporting EEG findings.

## Important
This is a cohort-discrimination task, **not** a clinical diagnosis. Your output identifies which research cohort the recording belongs to in this dataset; it does not claim that EEG alone can diagnose the underlying clinical condition.
\end{lstlisting}
\newpage

\begin{lstlisting}[style=inboxcompact]
== T5 Seizure Detection ==

## Clinical Context
continuous scalp EEG from a long-term epilepsy-monitoring session (pediatric epilepsy unit), 26-channel bipolar longitudinal montage, sampling rate 256 Hz, recording duration approximately 60 minutes.

## EEG Record
- **Record ID**: <record_id>
- **Duration Sec**: <duration_sec>
- **Montage**: bipolar_longitudinal
- **N Channels**: 26
- **Recording Type**: scalp
- **Sampling Rate**: 256.0

## Question
Search the entire recording for electrographic seizures. Report each seizure as an onset/offset interval in seconds from the start of the recording. If there is no seizure, return an empty list.
\end{lstlisting}
\begin{lstlisting}[style=inboxcompact]
== T6 Sleep Staging ==

## Clinical Context
33-year-old female, whole-night polysomnography (Sleep-EDFx, Sleep Cassette in-home, night 1), recording duration ~12.6 hours, 1508 epochs of 30 s each (AASM 5-class scoring required).

## EEG Record
- **Record ID**: <record_id>
- **Age**: <age>
- **Channels**: <n_channels>
- **Duration Sec**: <duration_sec>
- **Eeg Channels**: ['EEG Fpz-Cz', 'EEG Pz-Oz']
- **Epoch Sec**: 30
- **N Channels**: 7
- **N Epochs Total**: <n_epochs_total>
- **N Samples**: <n_samples>
- **Night**: <night>
- **Sampling Rate**: 100.0
- **Sex**: F
- **Source**: <source>

## Question
Score this whole-night PSG in 30-second epochs using AASM 5-class labels (W / N1 / N2 / N3 / REM).  Cover all 1508 epochs with a run-length-encoded hypnogram, and report REM_pct, N3_pct, and sleep_efficiency for the full recording.
\end{lstlisting}

\newpage
\paragraph{Long-horizon case studies (T5, T6).} We now present two long-horizon case studies in detail: a reasoning-and-tool-use trace followed by the corresponding outcome visualization for each task. The T5 case is a partial miss that illustrates a typical long-horizon failure; the T6 case illustrates how the agent turns multi-resolution whole-night features into a complete run-length-encoded hypnogram. All interaction rounds and tool calls are retained, while long thoughts and per-window tool arrays are abridged for readability.

\subparagraph{T5: Seizure detection trace.} The trace below shows how GPT-5.6-sol searches a one-hour recording for seizures. A full-hour screen at 10~s resolution exceeds the tool output budget, so the model falls back to a 15~s coarse screen. It then re-examines candidate periods at 5~s and 2~s resolution and confirms two seizures from a left-temporal theta evolution. It treats an early high-amplitude stretch as movement rather than a seizure, and therefore misses the first reference seizure. Long tool results are abridged.
\begin{lstlisting}[style=traj]
INPUT: ~1 h scalp EEG, 23-channel bipolar, 256 Hz. Report all seizure intervals (or none).
GROUND TRUTH (hidden): 3 seizures at [834,894], [2378,2497], [3362,3425] s

---- Round 1 ----
[Thinking] I'll first identify usable montage channels, then screen the full hour with coarse windowed rhythmicity/amplitude features and inspect any candidate periods at finer temporal resolution for sustained  ...
[Tool] get_channel_list
[Result] {"record_id": "EEG_0446", "n_channels": 38, "channels": [{"name": "FP1-F7", "type": "eeg", "hemisphere": "unknown", "region": "unknown", "marked_bad_i ...
---- Round 2 ----
[Thinking] I'll screen representative left- and right-temporal derivations in nonoverlapping 10-second windows; clinically significant focal seizures in this montage should usually recruit at least one temporal  ...
[Tool] compute_windowed_features
[Result] {"record_id": "EEG_0446", "error": "Output too large: this call would return 1 channels x 360 windows = 360 rows (~9645 tokens), exceeding the 8000-to ...
---- Round 3 ----
[Thinking] The 10-second full-hour request exceeded the response cap, so I'll use 15-second windows for the initial screen and then refine suspicious epochs with shorter windows.
[Tool] compute_windowed_features
[Result] {"record_id": "EEG_0446", "window_sec": 15, "step_sec": 15, "time_range": [0, 3600.0], "n_windows": 240, "features_computed": ["beta_relative", "rms_u ...
---- Round 4 ----
[Thinking] The left-temporal screen shows two sustained theta-enriched candidate periods (approximately 2370-2490 s and 3375-3420 s), amid several broad high-amplitude/low-beta stretches more suggestive of movem ...
[Tool] compute_windowed_features
[Result] {"record_id": "EEG_0446", "window_sec": 15, "step_sec": 15, "time_range": [0, 3600.0], "n_windows": 240, "features_computed": ["beta_relative", "rms_u ...
---- Round 5 ----
[Thinking] The first candidate is clearly left-predominant rather than bilaterally symmetric: left F7-T7 theta rises progressively across 2370-2490 s, while right F8-T8 remains relatively beta-rich except for li ...
[Tool] compute_windowed_features
[Result] {"record_id": "EEG_0446", "window_sec": 5, "step_sec": 5, "time_range": [2340, 2520.0], "n_windows": 36, "features_computed": ["alpha_relative", "beta ...
---- Round 6 ----
[Thinking] This segment demonstrates sustained left temporal evolution: T7-P7 amplitude increases from ~12-16 uV to >40 uV, theta becomes dominant, activity spreads posteriorly, and the spectral composition shif ...
[Tool] compute_windowed_features
[Result] {"record_id": "EEG_0446", "window_sec": 5, "step_sec": 5, "time_range": [3330, 3460.0], "n_windows": 26, "features_computed": ["alpha_relative", "beta ...
---- Round 7 ----
[Thinking] The later event has the same left mid-temporal signature-rapid theta dominance in T7-P7, posterior recruitment and amplitude growth, minimal right-temporal involvement, then abrupt attenuation after ~ ...
[Tool] compute_windowed_features
[Result] {"record_id": "EEG_0446", "window_sec": 2, "step_sec": 2, "time_range": [2320, 2520.0], "n_windows": 100, "features_computed": ["alpha_relative", "bet ...
---- Round 8 ----
[Thinking] For the first seizure, the earliest sustained change begins around 2354 s (theta becomes dominant in T7-P7 and remains organized while amplitude subsequently builds); termination is at ~2500 s, follow ...
[Tool] compute_windowed_features
[Result] {"record_id": "EEG_0446", "window_sec": 2, "step_sec": 2, "time_range": [3340, 3450.0], "n_windows": 55, "features_computed": ["alpha_relative", "beta ...

FINAL ANSWER: seizures = [2354,2500], [3358,3426]  (found #2 and #3; MISSED #1)
SCORE: Event-F1 0.80, Dice 0.798, 0 false alarms
\end{lstlisting}

\begin{figure}[H]
\centering
\includegraphics[width=\columnwidth]{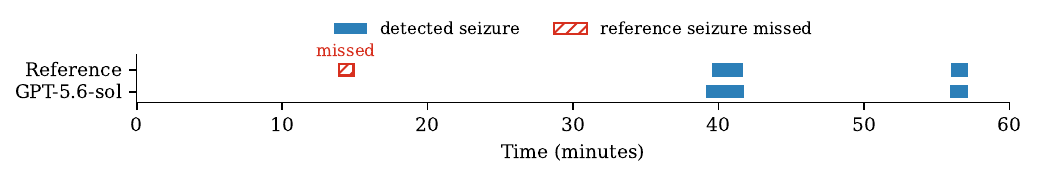}
\caption{T5 case (GPT-5.6-sol, CHB-MIT \tool{chb15_40}): reference and predicted seizure intervals over the one-hour recording. The model detects the second and third seizures but misses the first.}
\label{fig:t5_case}
\end{figure}

\subparagraph{T5: Seizure detection outcome.} Figure~\ref{fig:t5_case} visualizes the final structured output against the reference. GPT-5.6-sol detects the second and third seizures but misses the first, giving an Event-F1 of $0.80$ with no false alarms.

\subparagraph{T6: Sleep staging trace.} The trace below shows how GPT-5.6-sol analyzes the 12.6~h recording. It first checks channel quality, finds that the posterior EEG is affected by line noise and the submental EMG is flat, and therefore relies mainly on frontal EEG and horizontal EOG. After two requests exceed the tool-output budget, it progressively coarsens the analysis from 30-second to one-, two-, and three-minute windows. It then converts the inferred sleep boundaries and stage signatures into a run-length encoding covering all 1508 epochs.
\begin{lstlisting}[style=traj]
INPUT: 12.6 h whole-night PSG, 2 EEG + horizontal EOG + submental EMG, 1508 30-s epochs. Return a complete AASM 5-class RLE hypnogram and architecture fractions.
GROUND TRUTH (hidden): W=562, N1=109, N2=562, N3=105, REM=170 epochs

---- Round 1 ----
[Thinking] I will verify channel metadata and signal quality, then extract EEG/EOG/EMG features in manageable whole-night chunks before producing a contiguous RLE.
[Tool] get_channel_list(record_id="EEG_0003")
[Result] 7 channels: EEG Fpz-Cz, EEG Pz-Oz, EOG horizontal, Resp oro-nasal, EMG submental, Temp rectal, Event marker. The file header marks every channel as EEG.

---- Round 2 ----
[Thinking] Because the header mis-types the auxiliary channels, I will explicitly select the two EEG derivations and the named EOG/EMG channels, then check their quality.
[Tool] compute_noise_metrics(channels=["EEG Fpz-Cz","EEG Pz-Oz","EOG horizontal","EMG submental"], line_freq=50)
[Result] Fpz-Cz: RMS 23.2 uV, line-noise ratio 0.01. Pz-Oz: RMS 10.6 uV, line-noise ratio 0.32. EOG: RMS 67.1 uV. EMG: RMS 2.2 uV, flat-line ratio 1.0.

---- Round 3 ----
[Thinking] Frontal EEG is usable; posterior EEG has substantial line contamination; submental EMG is flat. I will rely on frontal EEG plus EOG and start with exact 30-s features for the first 300 epochs.
[Tool] compute_windowed_features(Fpz-Cz, features=[delta_relative,alpha_relative,sigma_power], window=30 s, step=30 s, time=0:9000 s)
[Result] ERROR: 300 rows would require about 8490 tokens, exceeding the 8000-token tool-output budget. Nothing was computed.

---- Round 4 ----
[Thinking] I will reduce the request to 240 epochs and add frontal RMS to distinguish high-amplitude slow-wave epochs from low-amplitude N1/REM-like epochs.
[Tool] compute_windowed_features(Fpz-Cz, features=[rms_uv,delta_relative,alpha_relative,sigma_power], window=30 s, step=30 s, time=0:7200 s)
[Result] 240 epoch-aligned rows returned. The trace remains high-delta/high-amplitude through about epoch 201, followed by an abrupt low-amplitude state change. [Per-window rows abridged.]

---- Round 5 ----
[Thinking] Exact 30-s coverage of all 1508 epochs is too large for the remaining tool budget. I will map the middle-night boundaries at 60-s resolution, then use stage-sensitive patterns to refine the RLE.
[Tool] compute_windowed_features(Fpz-Cz, features=[rms_uv,delta_relative,alpha_relative,sigma_power], window=60 s, step=60 s, time=7200:21600 s)
[Result] 240 one-minute rows returned. Repeated low-amplitude interruptions, sustained sigma-rich N2-like periods, and high-amplitude delta blocks are visible. [Per-window rows abridged.]

---- Round 6 ----
[Thinking] I will cover the remainder at two-minute resolution for broad architecture, then use EOG to separate REM-like low-amplitude bouts from wake.
[Tool] compute_windowed_features(Fpz-Cz, features=[rms_uv,delta_relative,alpha_relative,sigma_power], window=120 s, step=120 s, time=21600:45240 s)
[Result] 197 two-minute rows returned. The broad trace supports a main sleep interval ending near epoch 1188, followed by a prolonged high-amplitude post-sleep state. [Per-window rows abridged.]

---- Round 7 ----
[Thinking] I will quantify EOG amplitude over the full night at two-minute resolution; phasic EOG elevation during low-amplitude EEG should support REM candidates.
[Tool] compute_windowed_features(EOG horizontal, features=[rms_uv,zero_crossing_rate_hz], window=120 s, step=120 s, time=0:45240 s)
[Result] ERROR: 377 rows would require about 8214 tokens, exceeding the 8000-token tool-output budget. Nothing was computed.

---- Round 8 ----
[Thinking] I will use three-minute EOG windows. Combining the EEG and EOG traces, I infer wake at epochs 0-200 and 1188-1507, sleep onset through N1/N2 near epoch 201, early N3 near 286-346, and later N2/N3/REM cycles within the main sleep interval.
[Tool] compute_windowed_features(EOG horizontal, features=[rms_uv,zero_crossing_rate_hz], window=180 s, step=180 s, time=0:45180 s)
[Result] 251 three-minute EOG rows returned, providing whole-night ocular-activity context for the REM/wake decisions. [Per-window rows abridged.]

FINAL ANSWER:
hypnogram_rle = [
  [0,200,W], [201,216,N1], [217,228,N2], [229,235,N1],
  [236,245,N2], [246,251,N1], [252,285,N2], [286,346,N3],
  [347,351,N2], [352,359,REM], [360,364,N1], [365,383,N2],
  [384,388,W], [389,399,N1], [400,443,N2], [444,499,REM],
  [500,503,N1], [504,539,N2], [540,567,N3], [568,677,N2],
  [678,692,REM], [693,703,N2], [704,725,REM], [726,727,N1],
  [728,777,N2], [778,810,REM], [811,821,N1], [822,843,N2],
  [844,895,N3], [896,995,N2], [996,1049,REM], [1050,1157,N2],
  [1158,1187,REM], [1188,1507,W]
]
architecture = {REM_pct: 0.144562, N3_pct: 0.093501, sleep_efficiency: 0.651194}
SCORE: Macro-F1 0.770, Cohen's kappa 0.813
\end{lstlisting}

\subparagraph{T6: Sleep staging outcome.} Figure~\ref{fig:t6_case} shows the reference and predicted whole-night hypnograms. GPT-5.6-sol stages a 12.6~h night into 1508 epochs at a macro-F1 of $0.77$ and a Cohen's kappa of $0.81$.

\par\smallskip\noindent
\begin{minipage}{\columnwidth}
\centering
\includegraphics[width=\columnwidth]{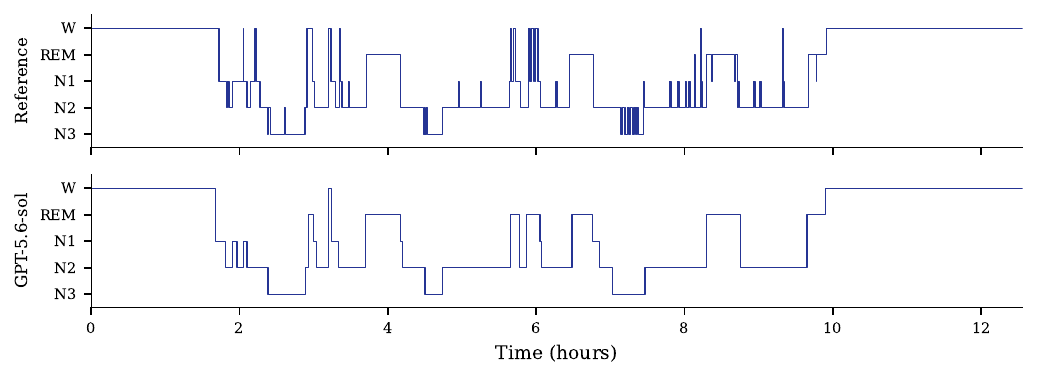}
\captionof{figure}{T6 case (GPT-5.6-sol, Sleep-EDFx \tool{SC4011E0}): reference and predicted whole-night hypnogram over 1508 thirty-second epochs (macro-F1 $0.770$, Cohen's kappa $0.813$).}
\label{fig:t6_case}
\end{minipage}
 
\end{document}